\documentclass[runningheads]{llncs}
\usepackage{graphicx}
\usepackage{comment}
\usepackage{amsmath,amssymb} 
\usepackage{color}
\usepackage{epsfig}
\usepackage{booktabs}
\usepackage{subfig}
\usepackage{multirow}
\usepackage{pifont}
\usepackage[symbol]{footmisc}

\newcommand{\cmark}{\ding{51}}%
\newcommand{\xmark}{\ding{55}}%

\definecolor{gray}{rgb}{0.5,0.5,0.5}
\definecolor{MyBlue}{rgb}{0,0,1.0}
\definecolor{MyPurple}{rgb}{0.4,0.1,0.5}
\definecolor{MyRed}{rgb}{0.8,0.2,0}
\definecolor{MyGreen}{rgb}{0,0.5,0.0}
\definecolor{MyGray}{rgb}{0.4,0.4,0.4}

\def\red#1{\textcolor{MyRed}{#1}}
\def\blue#1{\textcolor{MyBlue}{#1}}
\def\green#1{\textcolor{MyGreen}{#1}}
\def\purple#1{\textcolor{MyPurple}{#1}}

\def\first#1{\red{\textbf{#1}}}
\def\second#1{\blue{\textbf{#1}}}
\def\third#1{\purple{\textbf{#1}}}
\def\branch#1{\green{\textbf{#1}}}

\usepackage[width=122mm,left=12mm,paperwidth=146mm,height=193mm,top=12mm,paperheight=217mm]{geometry}

\begin{document}
\pagestyle{headings}
\mainmatter
\def\ECCVSubNumber{5243}  

\title{Regularized Adaptation for\\Stable and Efficient Continuous-Level Learning\\on Image Processing Networks} 

\titlerunning{Regularized Adaptation for Stable and Efficient Continuous-Level Learning}
%
\author{Hyeongmin Lee$^{\star1}$, Taeoh Kim$^{\star1}$,\\Hanbin Son$^1$, Sangwook Baek$^2$, Minsu Cheon$^2$, and Sangyoun Lee${\dagger^1}$}
\authorrunning{H. Lee, T. Kim, H. Son, S. Baek, M. Cheon, and S. Lee}
%
\institute{$^1$Yonsei University, Seoul, Korea \email{\{minimonia, kto, hbson, syleee\}@yonsei.ac.kr}\\
$^2$Samsung Research, Seoul, Korea \email{\{sw123.baek, minsu.cheon\}@samsung.com}}
\footnotetext[1]{Equal Contribution}
\footnotetext[2]{Corresponding Author}
\maketitle

\begin{abstract}
In Convolutional Neural Network (CNN) based image processing, most of the studies propose networks that are optimized for a single-level (or a single-objective); thus, they underperform on other levels and must be retrained for delivery of optimal performance. Using multiple models to cover multiple levels involves very high computational costs. To solve these problems, recent approaches train the networks on two different levels and propose their own interpolation methods to enable the arbitrary intermediate levels. 
However, many of them fail to adapt hard tasks or interpolate smoothly, or the others still require large memory and computational cost. In this paper, we propose a novel continuous-level learning framework using a Filter Transition Network (FTN) which is a non-linear module that easily adapts to new levels, and is regularized to prevent undesirable side-effects. Additionally, for stable learning of FTN, we newly propose a method to initialize non-linear CNNs with identity mappings. Furthermore, FTN is an extremely lightweight module since it is a data-independent module, which means it is not affected by the spatial resolution of the inputs.
Extensive results for various image processing tasks indicate that the performance of FTN is stable in terms of adaptation and interpolation, and comparable to that of the other heavy frameworks.


\end{abstract}

\begin{figure}
	\setlength{\belowcaptionskip}{-20pt}
	\begin{center}
		\subfloat[\small{Multi-task Learning}]
		{\includegraphics[width=0.495\linewidth]{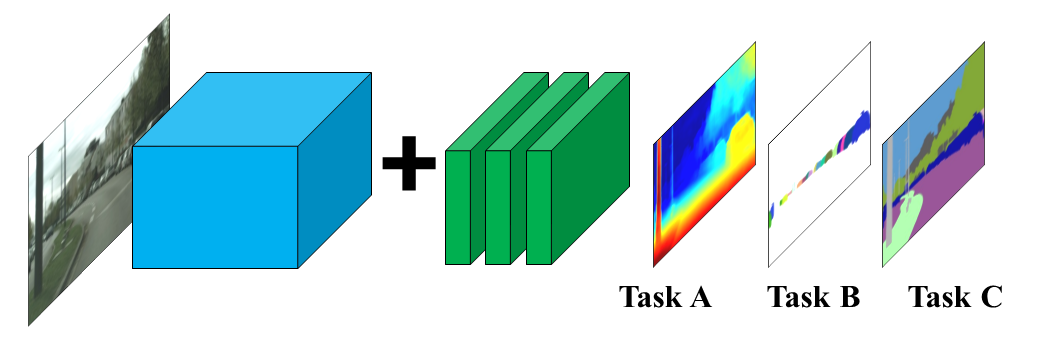}}
		\hfill
		\subfloat[\small{Multi-level Learning}]
		{\includegraphics[width=0.495\linewidth]{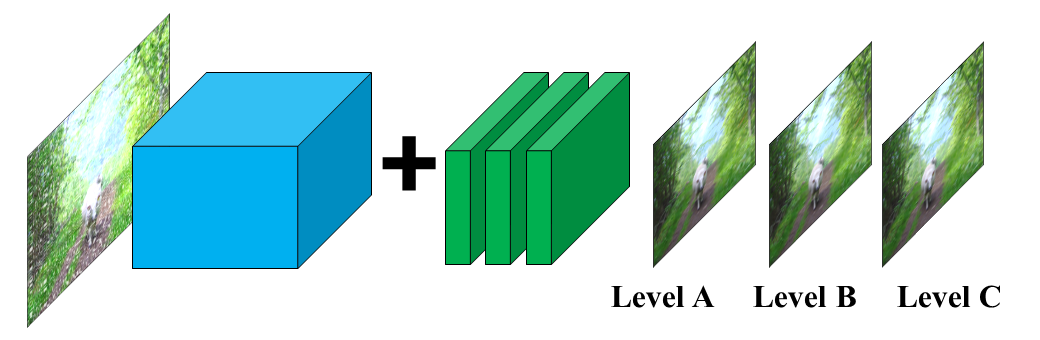}}\,\\[0.2ex]
		\subfloat[\small{Continuous-level Learning}]
		{\includegraphics[width=0.55\linewidth]{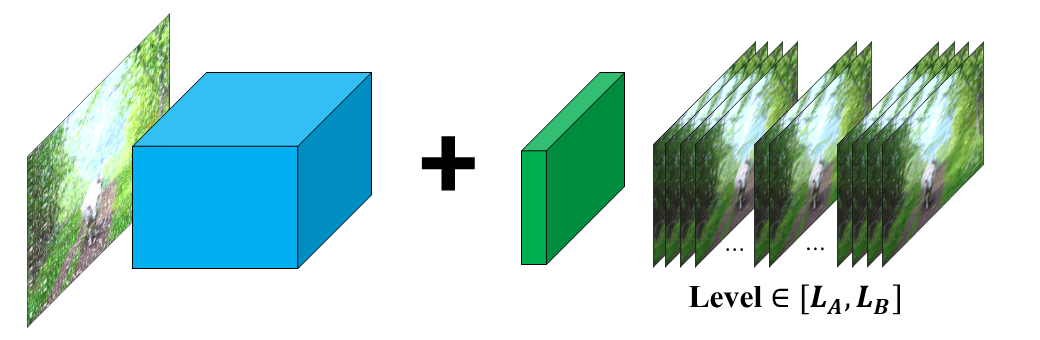}}\\[-1.3ex]
		\caption{\small{Comparison of multi-task learning, multi-level learning and continuous-level learning. Every task or level shares main network (\second{blue}) and introduces additional branch (\branch{green}) for task- or level-specific optimization}}
		\label{fig:cmll}
	\end{center}
\end{figure}

\section{Introduction}
\label{intro}

Image processing algorithms have various objectives that can include a combination of loss functions or a pair of target level-specific training datasets.
For example, in restoration task such as denoising, there is an optimized level for each input whose noise level is unknown; and in image synthesis, balancing between fidelity and naturalness \cite{pdt2018} depends on target applications.
In style transfer, the user hopes to control various styles and stylization strengths continuously.

However, most image processing deep networks are trained and optimized for a single-level. 
In this paper, the word \textbf{\textit{level}} can be one of the following examples: a target noise level (sigma of Gaussian or quality factor of JPEG), a combination of loss functions, or a strength of stylization.
If we want to handle $N$ multiple levels, we have to train $N$ different models or exploit the structure of multi-task learning \cite{edsr2017} (Fig. \ref{fig:cmll} (b)), which is very inefficient when $N$ increases.
In addition, in many image processing tasks, levels can be real numbers and each level produces semantically meaningful outputs.  
Therefore, to design a network for a continuous-level in an efficient way is a very practical issue. Fig. \ref{fig:cmll} describes the differences among multi-task learning, multi-level learning and continuous-level learning. Compared to multi-task learning, multi-level learning solves single-task multiple discrete-level problem. Continuous-Level Learning (CLL) is an extension of multi-level learning whose levels are continuous between two levels.

There have been several CLL frameworks~\cite{adafm2019,dynamicnet2019,cfsnet2019,dni2019,esrgan2018} and they commonly have following steps. In train phase, they train CNNs on the initial level, then the networks are fine-tuned or modified with tuning layer to the second level while some parameters are frozen. In test phase, they make their networks available at any intermediate level with their own interpolation methods. 
These steps are derived from the observations~\cite{adafm2019,dni2019}. The observation shows that the fine-tuned filters are similar to that of original filters which makes the interpolation space between filters linear approximately.

To achieve general, smooth and stable results, CLL algorithms have to satisfy three conditions. 
The first one is \textbf{\textit{good adaptation} (Sec. \ref{adaptation})}. 
After fine-tuning a network on the second level, since it contains parameters for both levels, the performance might be lower than the one trained only for the second level. Therefore, CLL frameworks have to be flexible so that they can adapt to the new levels well. 
The second one is \textbf{\textit{good interpolation} (Sec. \ref{interpolation})}. 
Even though the network works well on the two trained levels, it might not for the other intermediate levels. 
Therefore, it is important for the networks to maintain high performance and reasonable outputs for the intermediate levels. 
The last one is \textbf{\textit{efficiency} (Sec. \ref{efficiency})}. 
Since one of the main objective of CLL is to use a single network instead of using multiple networks trained for each level, requiring too large memory and computational resources is not practical for real-world applications.

\begin{table}[!t]
	\centering
	\caption{\small{Comparison of representative continuous-level learning methods. The word \textit{Regularized}} indicates it produces more smooth interpolation and less oversmoothing artifacts which will be discussed in Sec. \ref{regularization}. * indicates that it is achieved naturally from the linearity}
	\resizebox{0.99\linewidth}{!}{
					\begin{tabular}{lcccc}
				\toprule 
				& No Extra Memory &	Non-linear Adaptation &	Efficient	& Regularized \\
				\midrule
				AdaFM~\cite{adafm2019} & \cmark  & \xmark  & \cmark & \cmark*   \\
				DNI~\cite{dni2019,esrgan2018} & \xmark  &  \cmark  & \xmark  &  \xmark  \\
				Dynamic-Net~\cite{dynamicnet2019} & \cmark &  \cmark  & \xmark  &  \xmark \\
				CFSNet~\cite{cfsnet2019} & \cmark &  \cmark  & \xmark  &  \xmark \\
				FTN~(Ours) & \cmark  & \cmark  & \cmark & \cmark   \\
				\bottomrule
		\end{tabular}}%
	\label{tb:comparison}
\end{table}

Most of the prior approaches on CLL fail to satisfy all three conditions. AdaFM~\cite{adafm2019} introduces a tuning layer called feature modification layer for the second level which satisfies efficiency condition by just adding simple linear transition block (depth-wise convolution). However, the linearity reduces the flexibility of adaptation. Therefore, AdaFM cannot satisfy good adaptation condition, then it is not appropriate for more complex tasks such as style transfer. Deep Network Interpolation (DNI)~\cite{dni2019,esrgan2018} interpolates all parameters in two distinct networks trained for each level to increase flexibility. However, fine-tuning the network without any constraint cannot consider the initial level and it might lead to degraded performance on intermediate levels. Therefore, DNI fails to satisfy good interpolation condition. DNI also requires extra memory to save temporary network parameters and requires a third interpolated network for the inference. To satisfy both adaptation and interpolation condition, CFS-Net~\cite{cfsnet2019} and Dynamic~Net~\cite{dynamicnet2019} propose frameworks that interpolate the feature maps, not the model parameters using additional tuning branches. However, tuning branches require large memory and heavy computations up to double of the baseline networks. Therefore the efficiency condition is not satisfied. And these heterogeneous  networks can cause oversmoothing artifacts because each branch cannot consider the opposite-level. This side-effect will be discussed in Sec. \ref{interpolation}

In this paper, we propose a novel CLL method called \textit{Filter Transition Network (FTN)} that take convolutional neural network filters as input and learn the transitions between levels. Since FTN is a non-linear module, networks can be adapted to any new level. Therefore, it can cover general image processing tasks from simple Gaussian image denoising to complex stylization tasks. FTN \textit{transforms} the filters of the main network via other learnable networks, which makes the fine-tuning process regularized in stable parameter spaces for smooth and stable interpolation. Therefore, the good interpolation condition can be satisfied. 
For efficiency condition, from the observations in~\cite{adafm2019,dni2019}, FTN directly changes filters then it becomes data-agnostic. This prevents the increment of computational complexity, which is proportional to the spatial input resolution. Additionally, randomly initialized FTN makes the training process unstable since it directly changes model parameters. To solve this problem, we propose a method to initialize CNN modules with identity mapping. More specific comparisons with existing CLL frameworks are shown in Table~\ref{tb:comparison}.

In short, the proposed framework has following contributions:

\begin{itemize}
    \item We propose a novel CLL (Continuous-Level Learning) method which is not only flexible using non-linear transition but also regularized not to forget the initial level preventing side-effects.
    \item For the stability of learning for FTN, we propose a new initialization method that makes random non-linear convolutional network be identity.
    \item Our method is smooth and stable in terms of adaptation and interpolation, and significantly efficient in both memory and operations, while the performance is reasonable compared to the other competitive algorithms.
    \item We suggest an simple and efficient method for pixel-adaptive continous-level extensions without using pixel-adaptive convolutions.
\end{itemize}

\section{Related Work}

\noindent\textbf{Image Restoration.} CNN-based image restoration has shown great performance improvements over hand-crafted algorithms. After shallow networks, \cite{arcnn2015,srcnn2015}, some works stacked deeper layers, exploiting the advantages of residual skip-connection \cite{vdsr2016,dncnn2017}. Following the evolution of image recognition networks, restoration networks have focused on the coarse-to-fine scheme \cite{lapsrn2017}, dense connections \cite{rdn2018}, attentions \cite{rcan2018} and non-local networks \cite{nlrn2018}.
However, most networks are trained and optimized for a single level such as the Gaussian noise level in denoising, quality factor in JPEG compression artifact removal, and super-resolution scale in single-image super-resolution.
If the levels of training and test do not match, then optimal restoration performance cannot be achieved.
To solve the limitation, \cite{mildenhall2018burst,zhang2018ffdnet} proposed multiple noise-level training with a noise-level map, or noise estimation network \cite{guo2019toward} can be a solution. However, user cannot control at the test phase for better personalization~(\textit{e.g.} oversmoothing).

\noindent\textbf{The Perception-Distortion Trade-off.} In comparison with the general approach that attempts to reduce the pixel-error with the ground truth, some works \cite{argan2017,srgan2017,esrgan2018} attempted to produce more natural images using the generative power of GANs \cite{gan2014,cgan,dcgan}.
They used a combined loss of the fidelity term and adversarial term and then obtained better perceptual quality.
However, when a more adversarial loss is used, worse fidelity with the ground truth occurred due to the perception-distortion (PD) trade-off \cite{pdt2018}.
In \cite{pdt2018}, they proposed evaluating the restoration performance via a PD-plane \cite{pirm2018} considering the balance between fidelity and naturalness. However, the network must be retrained on another loss function to draw a continuous PD-function, which is a very time-consuming.

\noindent\textbf{Style Transfer.} With regard to image style transfer, Gatys~\emph{et~al.}~\cite{styletransfer2015} proposed a combination of content loss and style loss, and optimized content images via pre-trained feature extraction networks. Johnson~\emph{et~al.}~\cite{perceptual2016} made it possible to operate in a feed-forward manner using an image transformation network. However, a network trained on a single objective cannot control the balance between content and style and cannot handle continuous styles when it is trained on a single style.
Even though \cite{gatys2017controlling} can control several factors in the training phase and arbitrary (Zero-shot) style transfer such as \cite{huang2017arbitrary,sheng2018avatar} can handle infinite styles using adaptive instance normalization and style decorator, none of these can control continuous objectives (losses) at the test phase.

\section{Proposed Approach}
\subsection{Filter Transition Networks}
\label{defconv}

\begin{figure*}[!b]
	\setlength{\belowcaptionskip}{-5pt}
	\begin{center}
		\includegraphics[width=0.99\linewidth]{"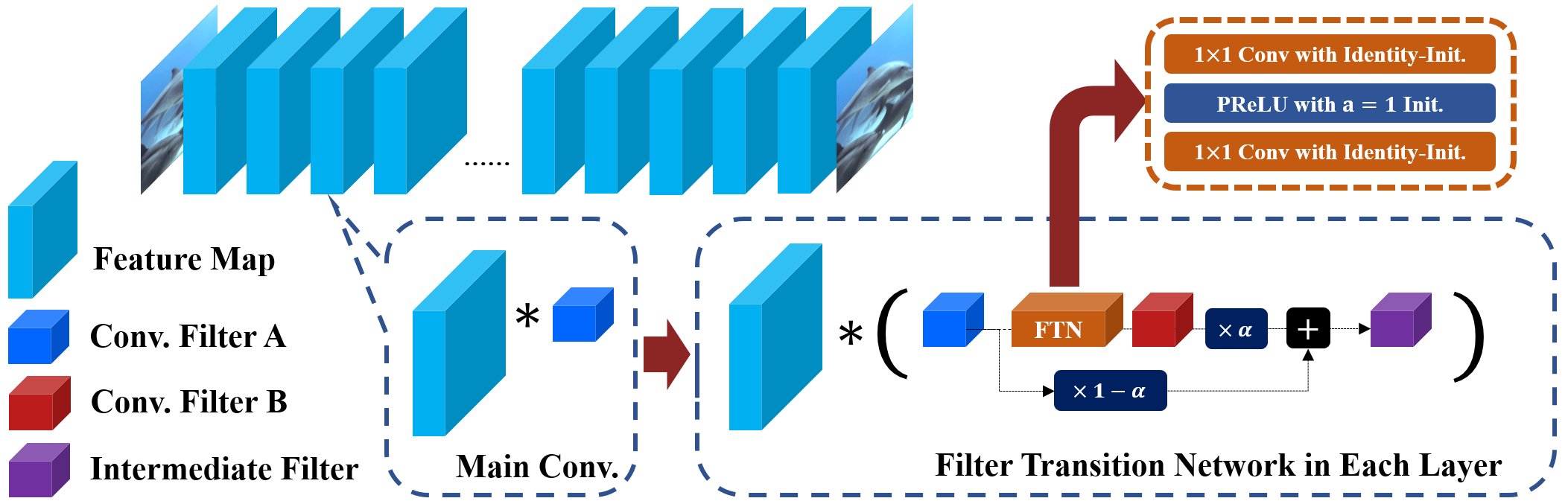}
	\end{center}
	\vspace*{-5mm}
	\caption{\small{Network architecture of our Filter Transition Network (FTN) when adapted in arbitrary main convolutional networks. Filter of main network (\second{blue}) is transformed via FTN for other levels (\first{red}). In inference phase, interpolated filter (\third{purple}) is used for intermediate levels}}
	\label{fig:network}
\end{figure*}

General concept of our module is same with the prior CLL frameworks; \textit{fine-tune and interpolate} which was described in Sec. \ref{intro}. Our overall framework is detailed in Fig. \ref{fig:network}. Our FTN module in an arbitrary convolutional layer can be described as
\begin{equation}
\textbf{X}_{i+1} = \textbf{X}_{i} * (\textbf{f}_{Ai} \times (1-\alpha) + FTN(\textbf{f}_{Ai}) \times \alpha )
\label{eq:ftn}
\end{equation}

\noindent where $\textbf{X}_{i}$ is an $i$-th convolutional feature map and $\textbf{f}_{Ai}$ is a corresponding filter. 

The FTN consists of two $1\times1$  convolutions with a $G$ grouped convolution  \cite{alexnet,resnext}, PReLU~\cite{prelu2015} activation functions, and skip-connection with weighted sum.
First, we train the main convolutional filter for the initial level with $\alpha=0$ which creates a vanilla convolution. Then, we freeze the main network and train the FTN only for the second level with $\alpha=1$, which breaks the skip-connection. Next, the FTN learn the task transition itself. To that end, the FTN approximates filters of the second level like by $FTN(\textbf{f}_{Ai}) \approx \textbf{f}_{Bi}$, where $\textbf{f}_{Bi}$ is an optimal filter for the second level. In the inference phase, we can interpolate between two filters (levels) by choosing $\alpha$ in the range of 0 to 1, and (\ref{eq:ftn}). Consequently, the FTN implicitly learns continuous transitions between levels, and $\alpha$ represents the amount of filter transition towards the second level.

The reasons why $1\times1$ convolution is used are two-fold: 1) it is lightweight and 2) padding is not required. Even if $1 \times 1$ filters cannot process spatially, they can be \textit{learned} considering spatial correlations because input of the FTN is very small (usually $3 \times 3 \times C$). Note that $1\times1$ filters in FTN convolves network filters, neither images nor features.

\subsection{Initialization of FTN}

When we train for the second-level, an ideal initialization of the FTN with an identity function is desired because of a convolution such as
\begin{equation}
\textbf{X}_{i+1} = \textbf{X}_{i} * (FTN(\textbf{f}_{Ai}))
\label{eq:ftn_init}
\end{equation}

\noindent However, networks are usually initialized with common methods such as \cite{xavier2010,prelu2015}, which will predict random filters from $\textbf{f}_{Ai}$. These kinds of initialization make the training very unstable unless a special trick is added. In our framework, every convolution and activation function is initialized as an identity function. Convolutions can easily become identities~\cite{adafm2019}. In activations, we use PReLU \cite{prelu2015} with an initial negative slope $a=1$, which will be learned through training. This initialization for non-linear layers makes the training more stable.

\begin{table}[!t]
	\centering
	\caption{\small{\textbf{Filter analysis for regularization.} Distance and Similarity between the two levels. We measure Mean Average Error (MAE) for linear filter interpolation and filter-wise normalized cosine similarity. The task is PD-controllable super-resolution and baseline network is \textbf{CFSNet-30}}}
	\resizebox{0.5\linewidth}{!}{
		\begin{tabular}{cc|ccc}
			\toprule
			&   & FTN (G=16)  & FTN  & Fine-tuning  \\ \midrule	
			& MAE  & \textbf{0.0082}  & 0.0118  & 0.0139  \\
			& Cos Sim.  & \textbf{0.9443} & 0.8937 & 0.8666  \\ \bottomrule	
	\end{tabular}}
	\label{tb:sr_similarity}
\end{table}

\subsection{Regularized Adaptation}
\label{regularization}

Good adaptation and good interpolation, which are mentioned in Section \ref{intro}, are in a trade-off relationship. For  good adaptation, the flexibility of the transition between the two levels is important. For example, AdaFM fails to adapt parameters between the levels which are in non-linear relationship, while producing good interpolation results due to its linearity. However, focusing on good adaptation without any constraint like DNI can make the network forget the initial level. It makes difficult to obtain smooth and meaningful intermediate filters through interpolation. In that sense, FTN is a regularized nonlinear method that satisfies both adaptation and interpolation conditions.

In our FTN, learnable transformation is shared across the spatial locations of filters and output channels in a layer. Only channel-wise features are used for adaptation. 
This can be viewed as a form of regularization and second-order representation of the filter. When group convolution ($G>2$) is used, feature extraction across channels is restricted, which results in stronger regularization. 
Table \ref{tb:sr_similarity} shows the filter distance with the filters of the main network when they are fine-tuned or passed through FTNs. The results show the filter-conditioned regularization is effective to prevent significant filter change. Performance of adaptation and interpolation will be discussed in the experiments sections.

\subsection{Complexity Analysis}
\label{efficiency}

\begin{table}[!t]
	\begin{center}
		\caption{\small{Overall computations, relative computations from baseline (in percentage), and number of parameters of the frameworks. Setting and network configurations are described in Sec. 4.1}}
		\resizebox{1.0\linewidth}{!}{
			\begin{tabular}{lcccccc}
				\toprule 
				Task & \multicolumn{2}{c}{Denoising} & \multicolumn{2}{c}{$\times$2 Super-Resolution} & \multicolumn{2}{c}{Style Transfer} \\
				\midrule
				Network & \multicolumn{2}{c}{AdaFM-Net} & \multicolumn{2}{c}{CFSNet-30} & \multicolumn{2}{c}{Transform-Net \cite{perceptual2016}} \\
				\cmidrule{2-7}          & GFLOPs  & Params(M)  & GFLOPs  & Params(M) & GFLOPs  & Params(M) \\
				\midrule
				Baseline & 25.11  & 1.41 & 155.96  & 2.37  & 40.42  & 1.68 \\
				+ CFSNet \cite{cfsnet2019} & 46.96 (87.02\%)  & 3.06 &  311.36 (99.64\%)  & 4.93  &  -  & -   \\
				+ AdaFM \cite{adafm2019} & 26.01 (3.58\%)  & \textbf{1.46}  & 162.50 (4.20\%)  & \textbf{2.47}  & 41.73 (3.24\%)  & \textbf{1.72} \\
				+ Dynamic-Net \cite{dynamicnet2019} & - & - & - & - & 62.24 (53.98\%) & 2.59 \\
				+ \textbf{FTN} & \textbf{25.36 (0.10\%)} & 1.83 & \textbf{156.34 (0.02\%)} & 3.01 & \textbf{40.86 (1.09\%)} & 2.06 \\
				+ \textbf{FTN(G=16)} & \textbf{25.13 (0.01\%)} & \textbf{1.44} & \textbf{156.00 (0.00\%)} & \textbf{2.41} & \textbf{40.46 (0.10\%)} & \textbf{1.70} \\
				\bottomrule
		\end{tabular}}%
		\label{tbl:complexity}
	\end{center}
\end{table}

Table \ref{tbl:complexity} compares the computational complexity and number of parameters with other frameworks. If any tuning layer with convolutions on a feature map is added, additional computations (MACs) are $H \times W \times K_{H} \times K_{W} \times C_{in} \times C_{out}$ where $H$, $W$, $K_{H}$, $K_{W}$, $C_{in}$ and $C_{out}$ are the height and width of the feature map (e.g., image size), height and width of the filter, and the number of input channels and output channels, respectively. Dominant computations arise from $H$ and $W$. In our network, which is a data-independent module, only $K_{H} \times K_{W} \times C_{in} \times (C_{out}/Groups) \times N$ is needed for a single tuning layer, where $N$ is the depth of FTNs. As shown in Table \ref{tbl:complexity}, FTNs have extremely reduced computational complexity and a similar or much lower number of parameters than other frameworks in various tasks and networks.

\section{Experiments}

\subsection{Experimental Settings}
\label{setup}

To understand recent CLL frameworks, we evaluate FTNs against fine-tuning (DNI) \cite{dni2019}, AdaFM \cite{adafm2019}, CFSNet \cite{cfsnet2019}, and Dynamic-Net \cite{dynamicnet2019} on four general image processing tasks. We add tuning layer of FTNs into every convolution, and same for AdaFM \cite{adafm2019} except the last layer to prevent boundary artifact. We add a ResBlock-wise (or DenseBlock-wise) tuning branch for CFSNet \cite{cfsnet2019}. For a fair comparison, the main networks are identical and shared across frameworks, and every hyper-parameter is identical except the tuning layers of each framework. More detailed configurations are described in supplementary material.

\noindent\textbf{Denoising \& DeJPEG.} We use two baseline networks that were proposed in \cite{adafm2019} and \cite{cfsnet2019}. The first network (AdaFM-Net \cite{adafm2019}) consists of 16 residual blocks as in \cite{edsr2017} with downsampling and upsampling layers. The second network (CFSNet-10 \cite{cfsnet2019}) consists of 10 residual blocks without downsampling or upsampling. We use DIV2K \cite{DIV2k} as the training set with a patch size of 48. We test on the CBSD68 \cite{cbsd68} dataset for denoising and LIVE1 \cite{live1} for deJPEG. We fine-tune the main network from the weaker noise (20 in denoising and 40 in deJPEG). Maximum PSNR is obtained via grid search of $\alpha$.

\noindent\textbf{PD-Controllable Super-resolution.} In image super-resolution, as reported in \cite{pdt2018}, fidelity and naturalness exhibit trade-offs. A comparison between algorithms should consider this trade-off by plotting perception (fidelity)-distortion (naturalness) curves. Drawing this curve is possible by changing weights between loss terms. As in \cite{esrgan2018}, we train phase 1 using L1 loss, and fine-tune using a  combined loss of L1, Perceptual (VGG, \cite{perceptual2016}) and GAN losses. We evaluate using two baseline networks that were proposed for \cite{cfsnet2019} (CFSNet-30) and \cite{esrgan2018} (ESRGAN). CFSNet-30 is the deeper version of CFSNet-10 for image super-resolution, and ESRGAN consists of multiple densely connected \cite{rdn2018} residual blocks. We use DIV2K as the training set with patch size a 128 and PIRM \cite{pirm2018} as the test set. PSNR and SSIM \cite{ssim} are used as distortion metrics, and NIQE \cite{niqe} and the Perceptual Index \cite{pirm2018} are used as perception metrics.

\noindent\textbf{Style Transfer.} In style transfer, we use Transform-Net which was proposed in \cite{perceptual2016} with instance normalization \cite{instance_norm}. We follow the settings of Dynamic-Net \cite{dynamicnet2019}. COCO 2014 train dataset \cite{cocodataset} is used for training. From the main network, Dynamic-Net inserts three tuning branches into pre-defined layers, while FTNs are inserted in every convolution layer. This means that FTNs have more opportunities to control in a layer-wise manner (See Fig. 1 of supplementary material of Dynamic-Net \cite{dynamicnet2019}).

\begin{table}[!t]
	\centering
	\caption{\small{\textbf{Ablation study for structures of FTN.} Average PSNR (dB) on CBSD68 denoising test dataset. Unseen noise levels are denoted with 
	*. The baseline network is \textbf{AdaFM-Net}. We color the \first{best} and the \second{second best}}}
	\setlength{\belowcaptionskip}{-10pt}
	\resizebox{0.4\linewidth}{!}{
		\begin{tabular}{cc|ccccc}
			\toprule
			& Noise Level $\sigma$ & 20  & 30*  & 40*  & 50  \\ \midrule	
			& From Scratch  & 32.44  & 30.37  & 29.00  & 27.96  \\
			& FTN  & \textbf{32.44}  & 30.18  & 28.90  & \first{28.04} \\
			& FTN-gc4  & \textbf{32.44}  & \second{30.31}  & \second{28.92}  & 28.02  \\
			& FTN-gc16  & \textbf{32.44}  & \first{30.37}  & \first{28.99}  & 28.01   \\
			& FTN-deeper & \textbf{32.44}  & 30.06  & 28.81  & \second{28.03} \\
			& FTN-spatial & \textbf{32.44}  & 30.16  & 28.88  & \first{28.04} \\ \bottomrule	
	\end{tabular}}
	\label{tb:dn_ablation}
\end{table}

\begin{figure}[!t]
    \setlength{\abovecaptionskip}{-9pt}
    \setlength{\belowcaptionskip}{-9pt}
	\begin{center}
		\includegraphics[width=0.58\linewidth]{"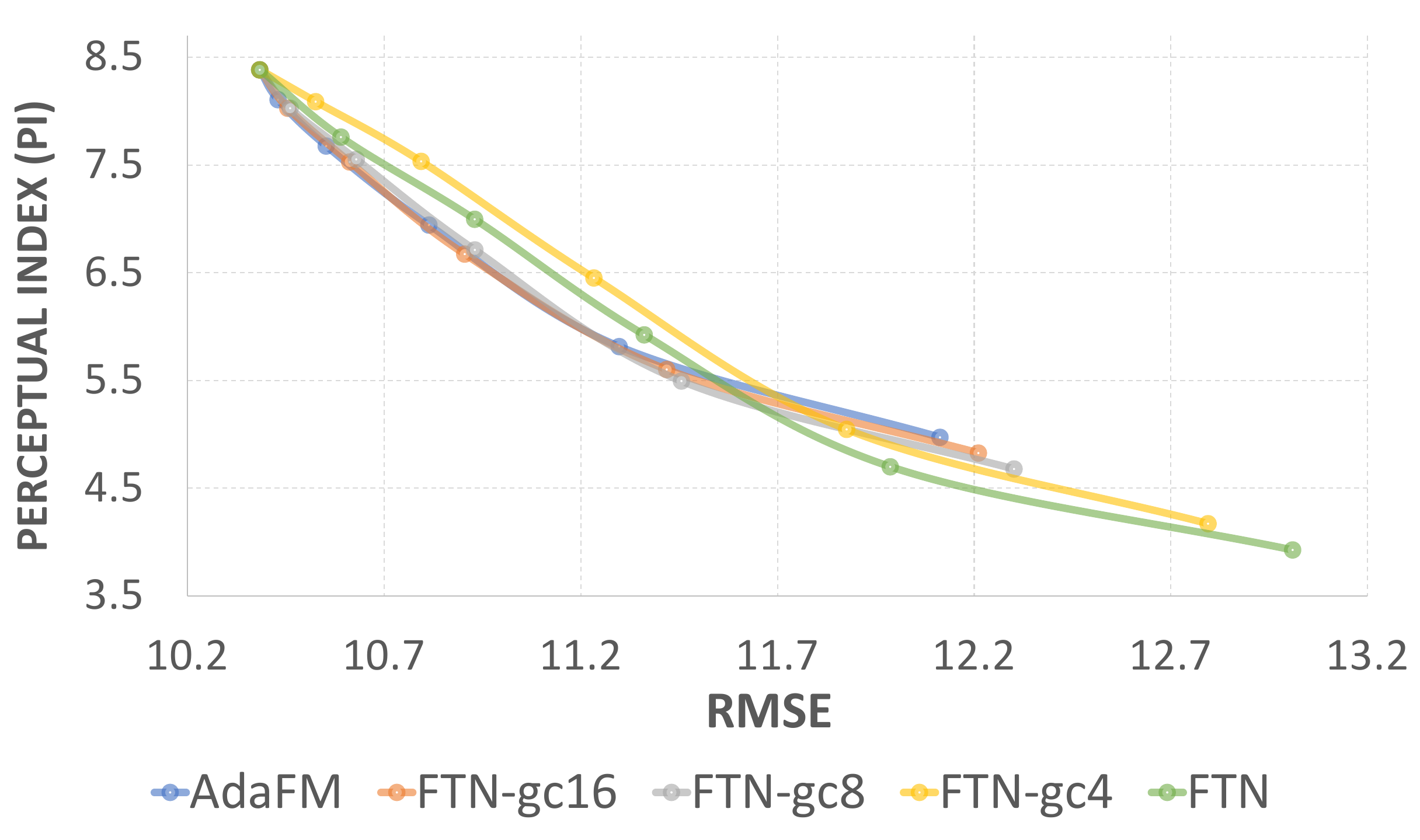}
	\end{center}
	\caption{\small{\textbf{Ablation study for linearities of transition modules in PD-control}. Result show that linear module cannot transit well toward the second-level}}
	\label{fig:sr_abl}	
\end{figure}

\subsection{Ablation Study}
\label{ablation}

First, to check the effect of regularization, we perform an ablation study on AdaFM-Net to compare different structures of FTNs in Table \ref{tb:dn_ablation}. We define various versions of FTN: more regularized~(FTN-\textit{gc}) or less regularized~(FTN-\textit{deeper}, FTN-\textit{spatial}). FTN-\textit{gc} is the version that the convolution layers are replaced by group convolution. FTN-\textit{deeper} is a three-layer version of the FTN whose intermediate results are worse than others because too much modification hurts the interpolation results. FTN-\textit{spatial} is a depth-wise convolution version whose performance is inferior to other channel-wise convolutions. In Table \ref{tb:dn_ablation}, the versions with less regularization shows better adaptation performance, and the ones with more regularization shows better interpolation performance.

Also, Fig.~\ref{fig:sr_abl} is the comparison over the number of groups in FTN-\textit{gc} on the PD-controllable super-resolution. The curves also prove that stronger regularization improves the interpolation performance, while makes further adaptation difficult.

\subsection{Adaptation Performance}
\label{adaptation}

\begin{table}[!t]
	\centering
	\caption{\small{\textbf{Gaussian Denoising Results.} Average PSNR (dB) on CBSD68 test dataset. Unseen noise levels are denoted with *. We color the \first{best} and the \second{second best}}}
	\resizebox{1.0\linewidth}{!}{
		\begin{tabular}{c|cccc|cccc|cccc|cccc}
			\toprule
			\multicolumn{1}{c|}{} & \multicolumn{8}{c|}{\large{AdaFM-Net}} & \multicolumn{8}{c}{\large{CFSNet-10}}  \\ 
			\multicolumn{1}{c|}{} & \multicolumn{4}{c|}{Short Adaptation} & \multicolumn{4}{c|}{Long Adaptation} & \multicolumn{4}{c|}{Short Adaptation} & \multicolumn{4}{c}{Long Adaptation} \\ 
			\midrule 
			
			Noise Level & 20  &  40*   & 60*  & 80	& 20   & 40*   & 60*  & 80  & 20  &  40*   & 60*  & 80 & 20  &  40*   & 60*  & 80 \\
			\midrule
			
			DNI \cite{dni2019} & \textbf{32.44} & 28.20 & \second{26.98} & 25.97 & \textbf{32.44}  & 27.54 & 26.74  & 25.93 & \textbf{32.42} & \first{28.87}  & \first{27.01} & \first{25.96} & \textbf{32.42}   & \first{28.89} & \first{27.11}  & \first{25.95} \\
			AdaFM \cite{adafm2019} & \textbf{32.44} & 28.17  & 26.77& 25.96 & \textbf{32.44}  & 28.28 & 26.80  & 25.96 & \textbf{32.42}  & 28.48  & 26.75 & 25.84 & \textbf{32.42}  & 28.42 & 26.72  & 25.82 \\
			CFSNet \cite{cfsnet2019} & \textbf{32.44}  & 28.41  & 26.87 & \second{26.00} & \textbf{32.44}  & 28.44 & 26.86  & \second{26.00} & \textbf{32.42} & 28.65  & 26.95 & \second{25.93} & \textbf{32.42}  & 28.67 & 26.98  & \second{25.93} \\
			\textbf{FTN-gc16} & \textbf{32.44}  & \first{28.78}  & \first{27.05} & 25.98  & \textbf{34.44}  & \second{28.78}  & \first{27.05} & 25.98 & \textbf{32.42}  & \second{28.77}  & \second{27.00} & 25.89 & \textbf{32.42}& \second{28.77} & \second{27.00}  & 25.88 \\
			\textbf{FTN-gc4} & \textbf{32.44}  & \second{28.64}  & 26.95 & \second{26.00} & \textbf{34.44} &  26.62 & \second{26.92}  & \second{26.00} & \textbf{32.42}  & 28.65  & 26.90& 25.90  & \textbf{32.42}  & 28.64 & 26.904  & 25.90 \\
			\textbf{FTN} & \textbf{32.44} & 28.48   & 26.89 & \first{26.03} & \textbf{34.44}  & \first{28.86}  & 26.79 & \first{26.03} & \textbf{32.42}  & 28.45 & 26.86 & 25.92 & \textbf{32.42}  & 28.49 & 26.89  & \second{25.93} \\ 
             \hline
	\end{tabular}}
	\label{tb:dn_result}
\end{table}

\begin{figure*}[!t]
	\setlength{\abovecaptionskip}{-1pt}
	\setlength{\belowcaptionskip}{-25pt}
	\begin{center}
		\captionsetup[subfigure]{labelformat=empty}
		\rotatebox{90}{\makebox[19mm][c]{\small{\textsc{Style A}}}}
		\subfloat
		{\includegraphics[height=0.157\linewidth]{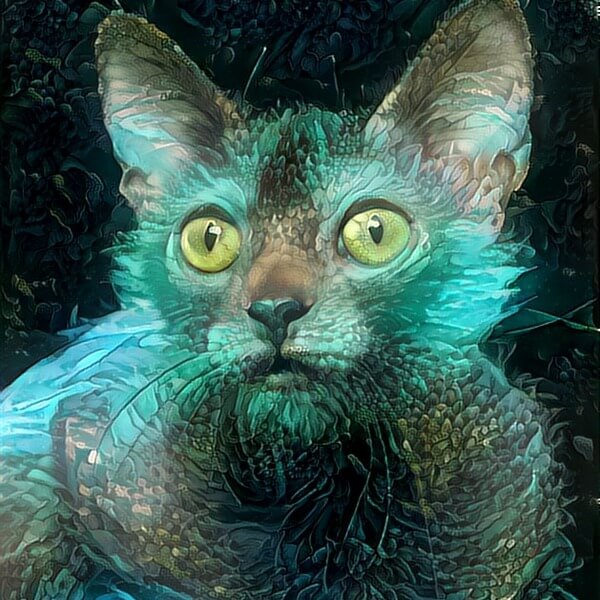}}\
		\hfill
		\rotatebox{90}{\makebox[19mm][c]{\small{\textsc{Content}}}}
		\subfloat
		{\includegraphics[height=0.157\linewidth]{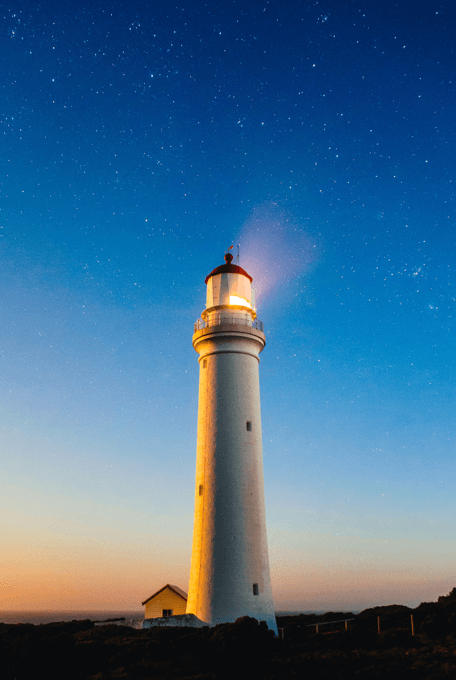}}\
		\hfill
		\rotatebox{90}{\makebox[19mm][c]{\small{\textsc{Style B}}}}
		\subfloat
		{\includegraphics[height=0.157\linewidth]{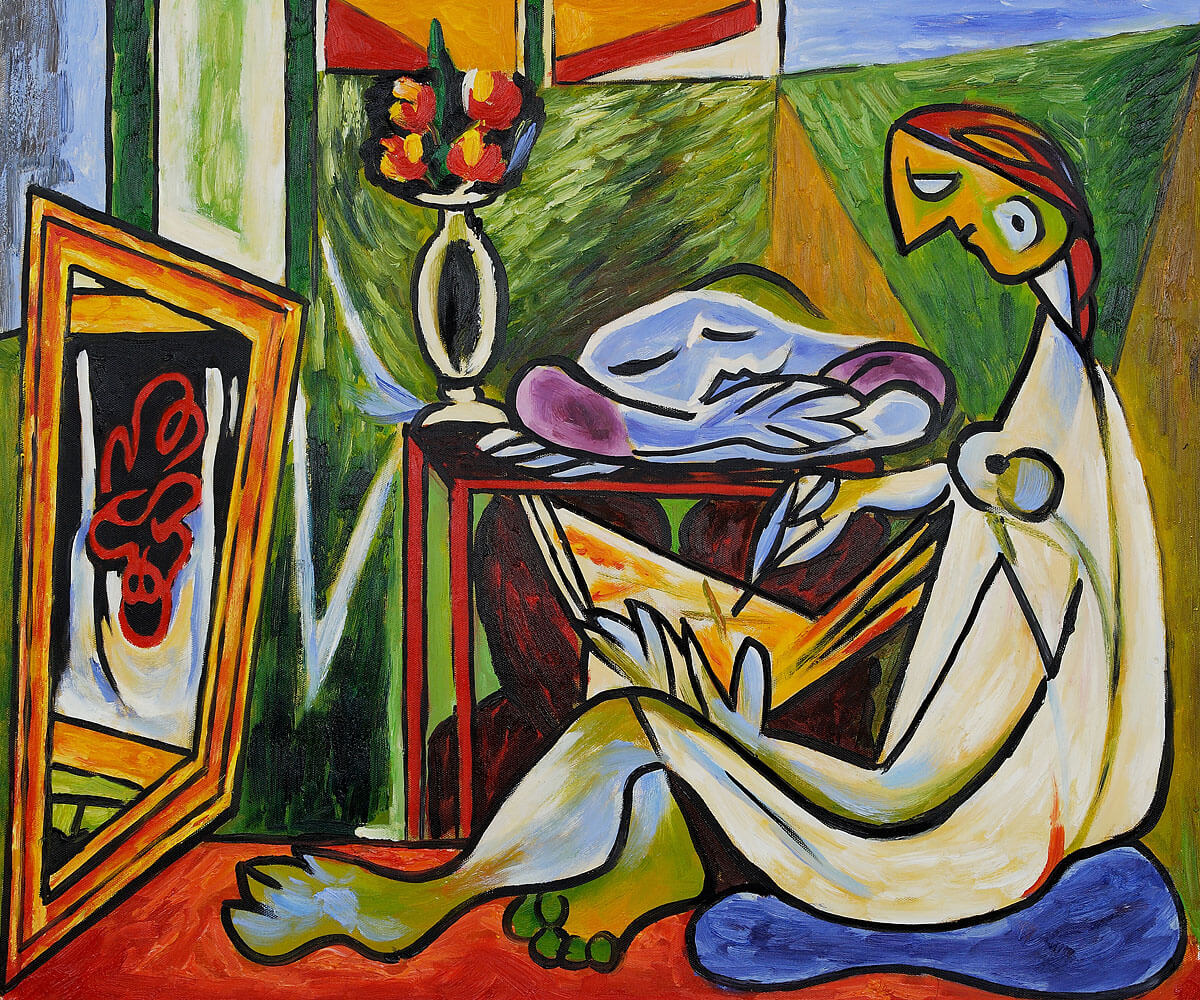}}\
		\\[-2ex]
		\rotatebox{90}{\makebox[27mm][c]{\small{\textsc{AdaFM}}}}
		\subfloat
		{\includegraphics[width=0.155\linewidth]{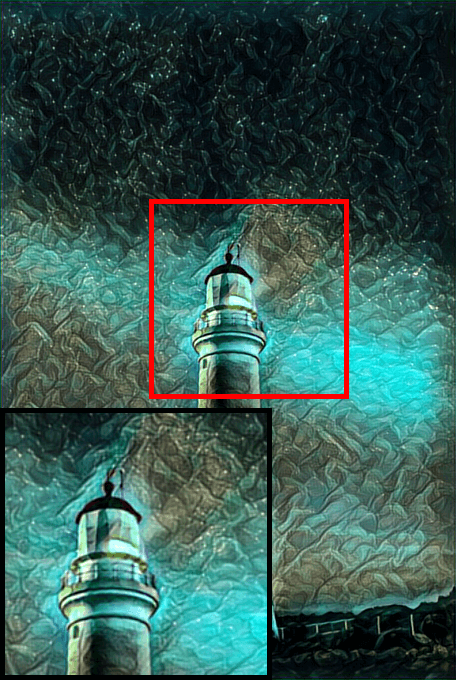}}\
		\hfill
		\subfloat
		{\includegraphics[width=0.155\linewidth]{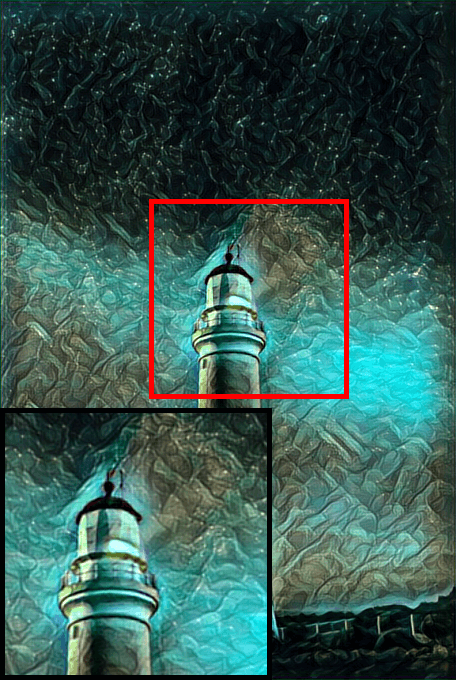}}\
		\hfill
		\subfloat
		{\includegraphics[width=0.155\linewidth]{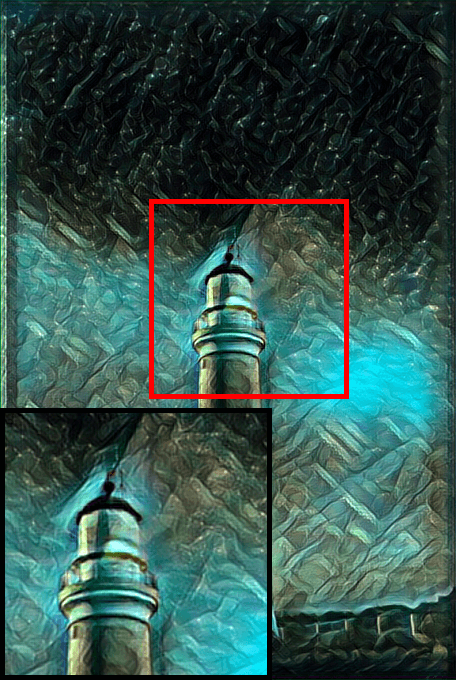}}\
		\hfill
		\subfloat
		{\includegraphics[width=0.155\linewidth]{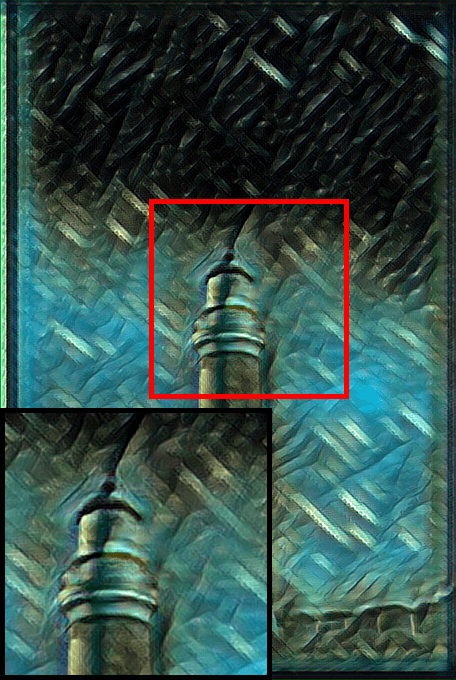}}\
		\hfill
		\subfloat
		{\includegraphics[width=0.155\linewidth]{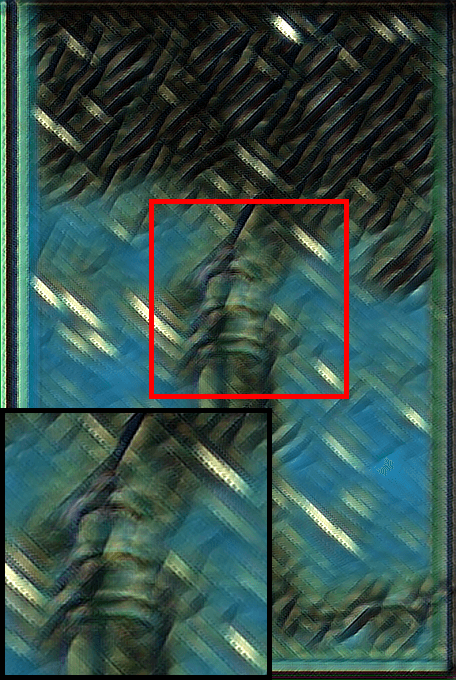}}\
		\hfill
		\subfloat
		{\includegraphics[width=0.155\linewidth]{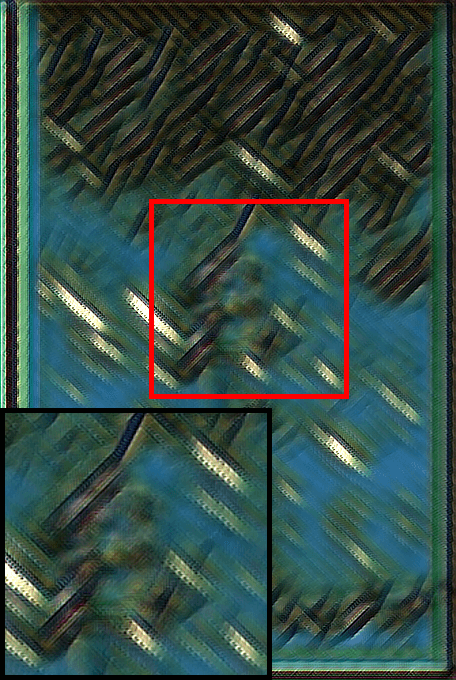}}\
		\\[-2.5ex]
		\rotatebox{90}{\makebox[27mm][c]{\small{\textsc{Dynamic-Net}}}}
		\subfloat
		{\includegraphics[width=0.155\linewidth]{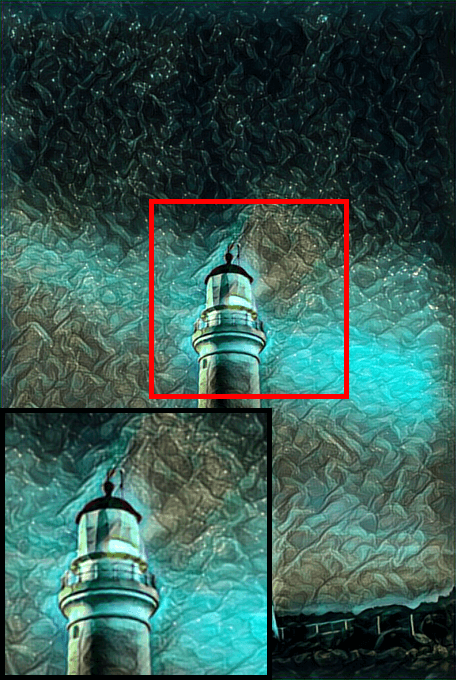}}\
		\hfill
		\subfloat
		{\includegraphics[width=0.155\linewidth]{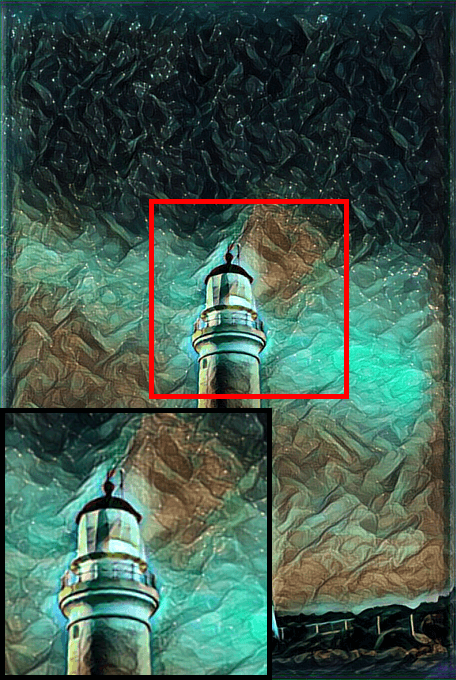}}\
		\hfill
		\subfloat
		{\includegraphics[width=0.155\linewidth]{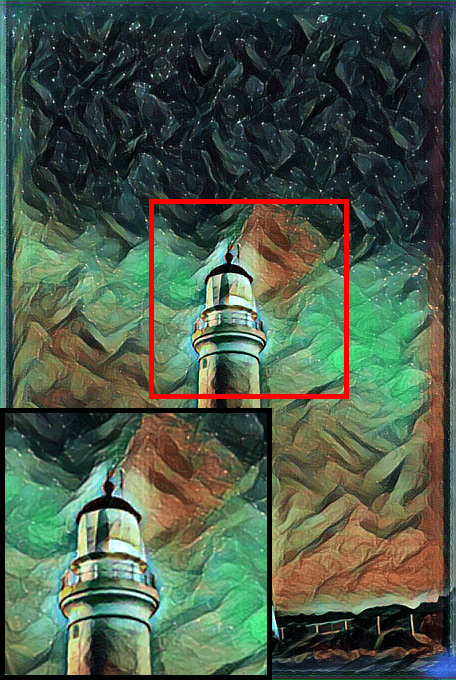}}\
		\hfill
		\subfloat
		{\includegraphics[width=0.155\linewidth]{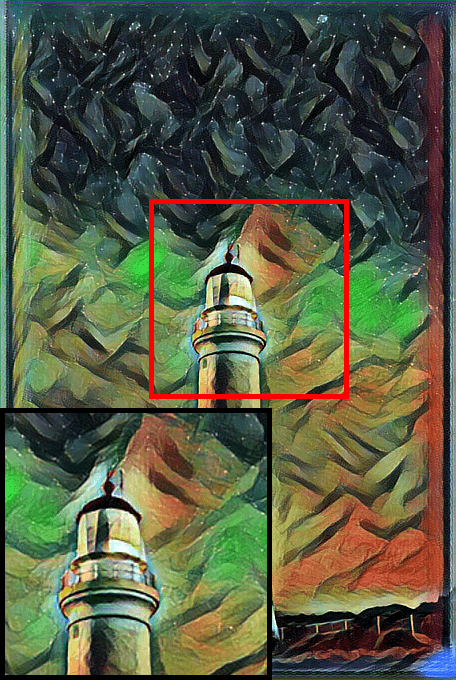}}\
		\hfill
		\subfloat
		{\includegraphics[width=0.155\linewidth]{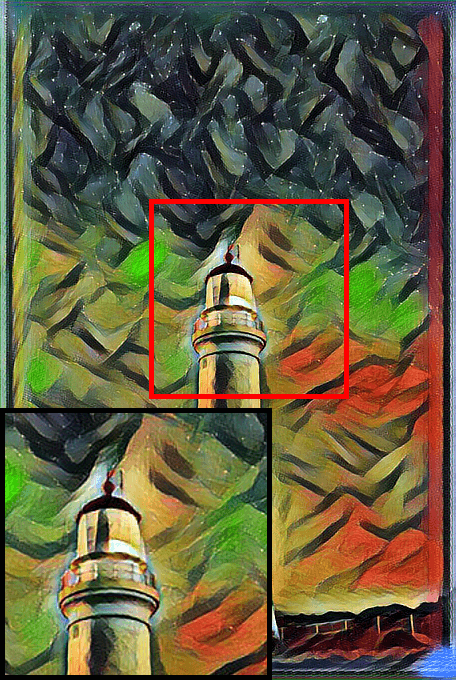}}\
		\hfill
		\subfloat
		{\includegraphics[width=0.155\linewidth]{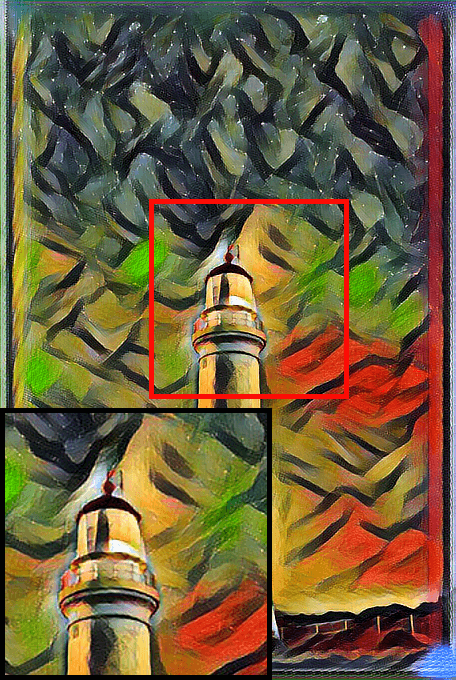}}\
		\\[-2.5ex]
		\rotatebox{90}{\makebox[27mm][c]{\small{\textsc{FTN}}}}
		\subfloat[$\alpha=0.0$]
		{\includegraphics[width=0.155\linewidth]{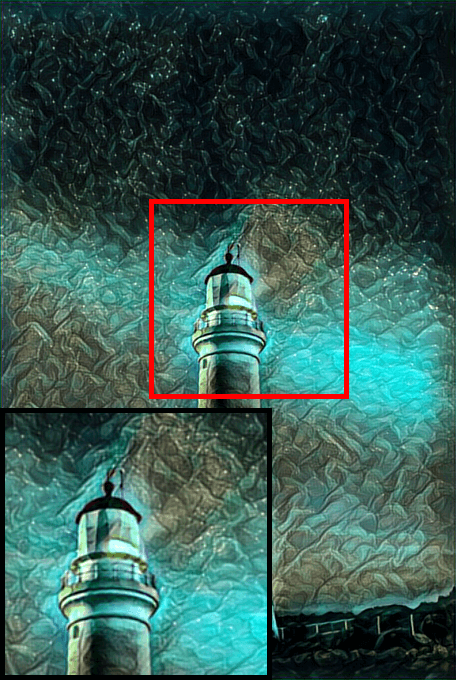}}\
		\hfill
		\subfloat[$\alpha=0.2$]
		{\includegraphics[width=0.155\linewidth]{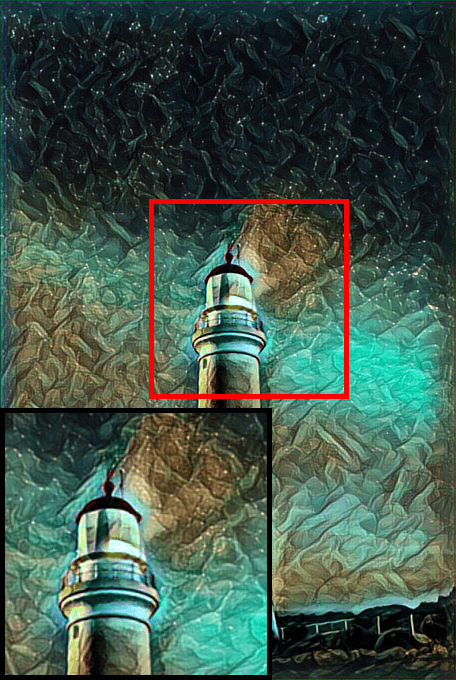}}\
		\hfill		
		\subfloat[$\alpha=0.4$]
		{\includegraphics[width=0.155\linewidth]{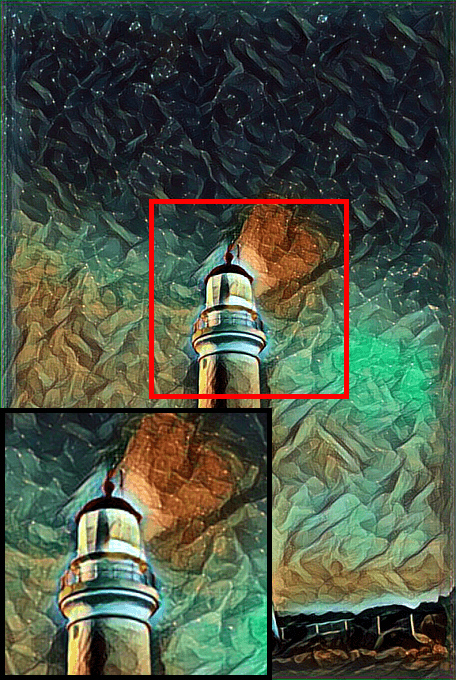}}\
		\hfill
		\subfloat[$\alpha=0.6$]
		{\includegraphics[width=0.155\linewidth]{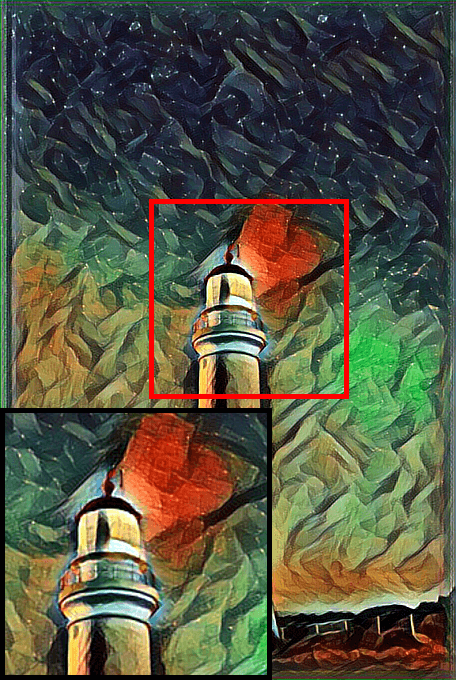}}\
		\hfill
		\subfloat[$\alpha=0.8$]
		{\includegraphics[width=0.155\linewidth]{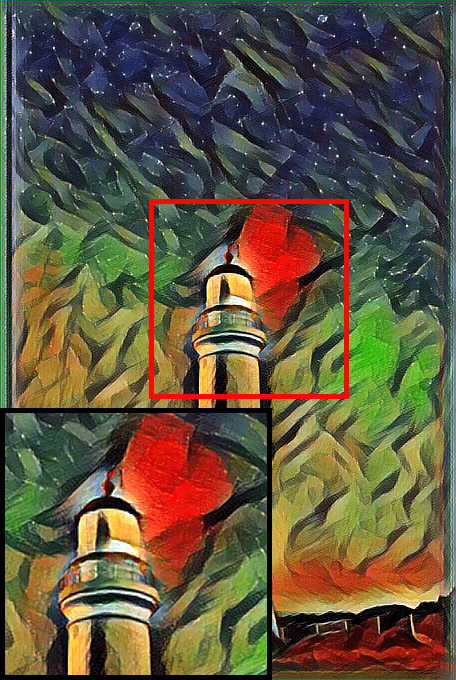}}\
		\hfill
		\subfloat[$\alpha=1.0$]
		{\includegraphics[width=0.155\linewidth]{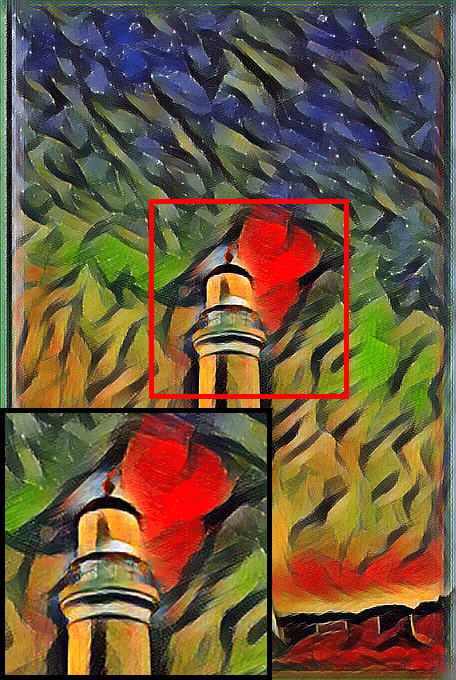}}\
		\caption{\small{\textbf{Visual comparison of controllable style transfer results between two styles.} Results show that linear module cannot transit semantic task. Compared to Dynamic-Net, FTN shows better adaptation results}}
		\label{fig:style_visual}
	\end{center}
\end{figure*}

We compare the adaptation performances to the other CLL methods. The adaptation performance means the performance on the second level compared to a network only trained for the level. Table~\ref{tb:dn_result} shows the adaptation/interpolation performances on the denoising task (Results on deJPEG task and result images are reported in supplementary material.). In the table, the adaptation performance indicates PSNR in noise level 80. \textit{Short/Long adaptation} indicates the training time for the second-level adaptation. More adaptation results are described in supplementary material. The table shows that there is a little difference between adaptation performances over the compared methods, including even AdaFM that uses linear adaptation. This is because denoising and deJPEG tasks require their model parameters to be changed less as the level changes. On the other hand, according to Fig.~\ref{fig:sr_abl}, the adaptation of AdaFM is less than FTN-\textit{gc16}, which is the most regularized version of FTN. Besides, Fig.~\ref{fig:style_visual} shows the results of the other aspects. The figure is the result on style transfer task, which requires large transition of the model parameters as the style changes. According to Fig.~\ref{fig:style_visual}, AdaFM fails to adapt from the style A to the style B. These results show that the linear adaptation has limitation in reaching the hard second level. In Dynamic-Net, it cannot deliver the second style smoothly because it only changes three pre-defined layer while FTN changes all convolutional filters. More results for stylization are described in supplementary material.

\subsection{Interpolation Performance}
\label{interpolation}

Although a CLL algorithm successfully adapt its network to the second level, it might fail for the interpolated levels. Especially, when the network forgets the initial level (i.e., loses correlation with the initial status) during the adaptation process, the interpolated parameters cannot work for the intermediate levels anymore. We evaluate interpolation performance in following three aspects.

\begin{figure*}[!t]
	\setlength{\belowcaptionskip}{-17pt}
	\begin{center}
		\captionsetup[subfigure]{labelformat=empty}
		\rotatebox{90}{\makebox[40mm][c]{{\textsc{CFSNet-30}}}}
		\subfloat
		{\includegraphics[width=0.450\linewidth]{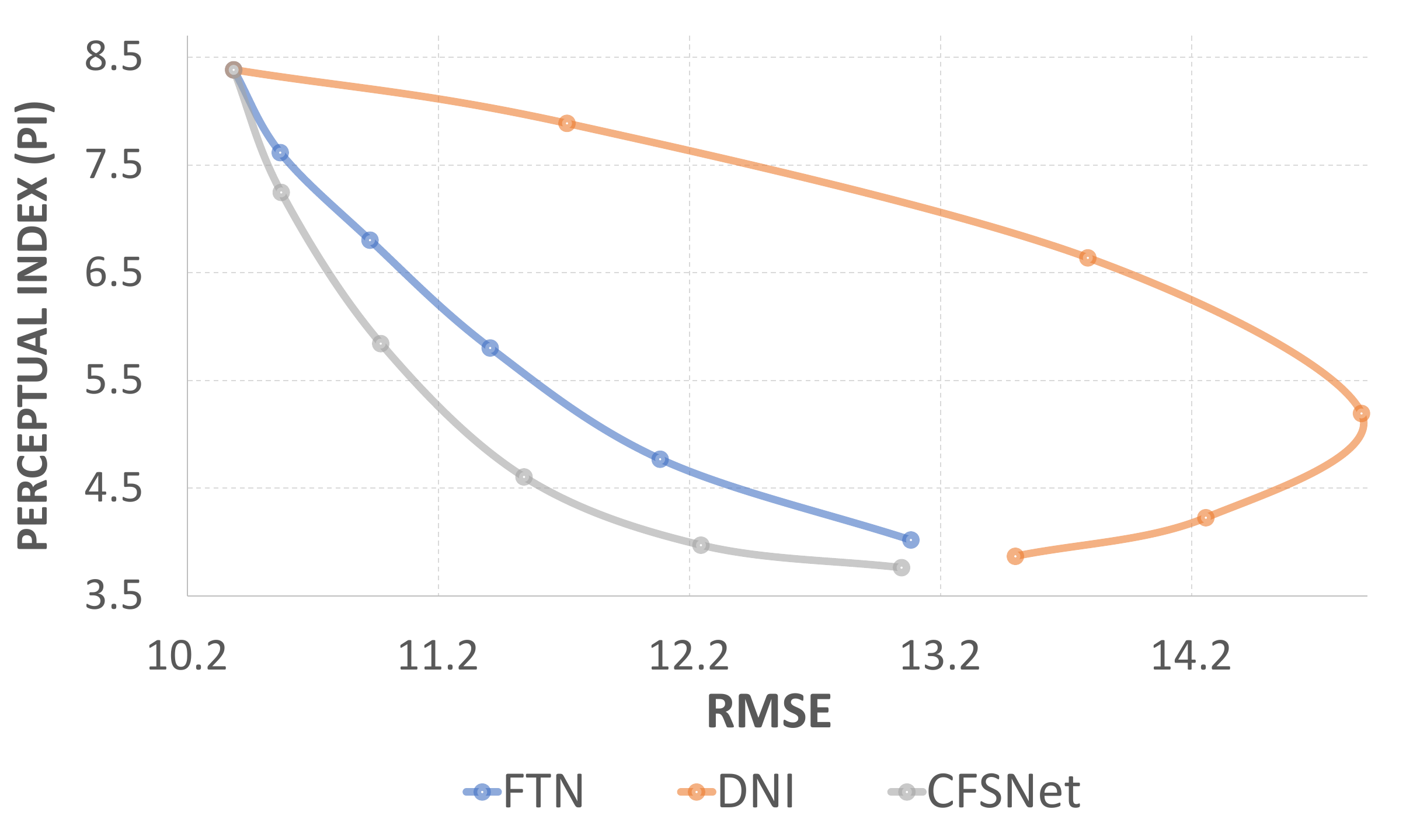}}\
		\subfloat
		{\includegraphics[width=0.450\linewidth]{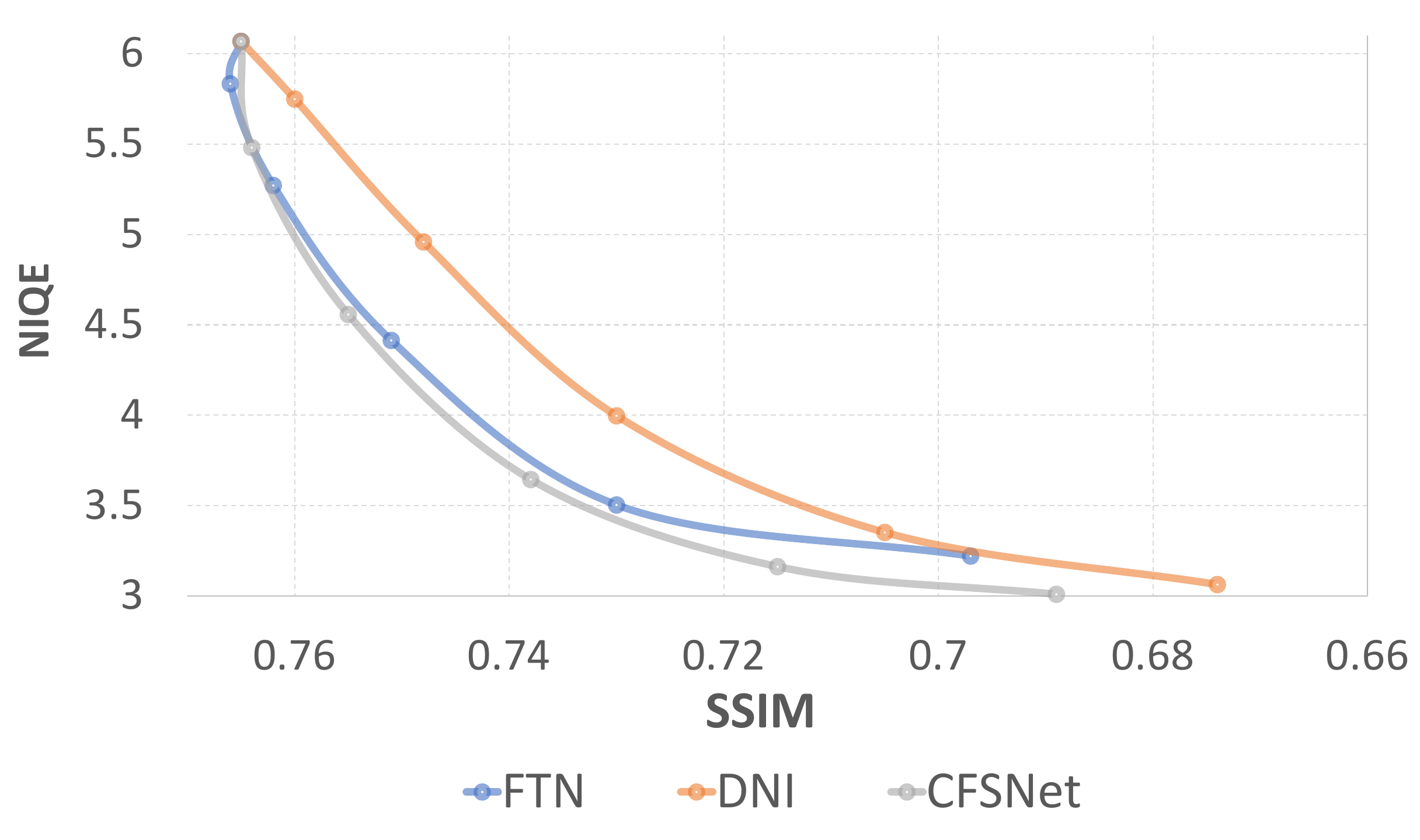}}\
		\\[-8ex]
		\rotatebox{90}{\makebox[40mm][c]{{\textsc{ESRGAN}}}}
		\subfloat[{RMSE-PI}]
		{\includegraphics[width=0.450\linewidth]{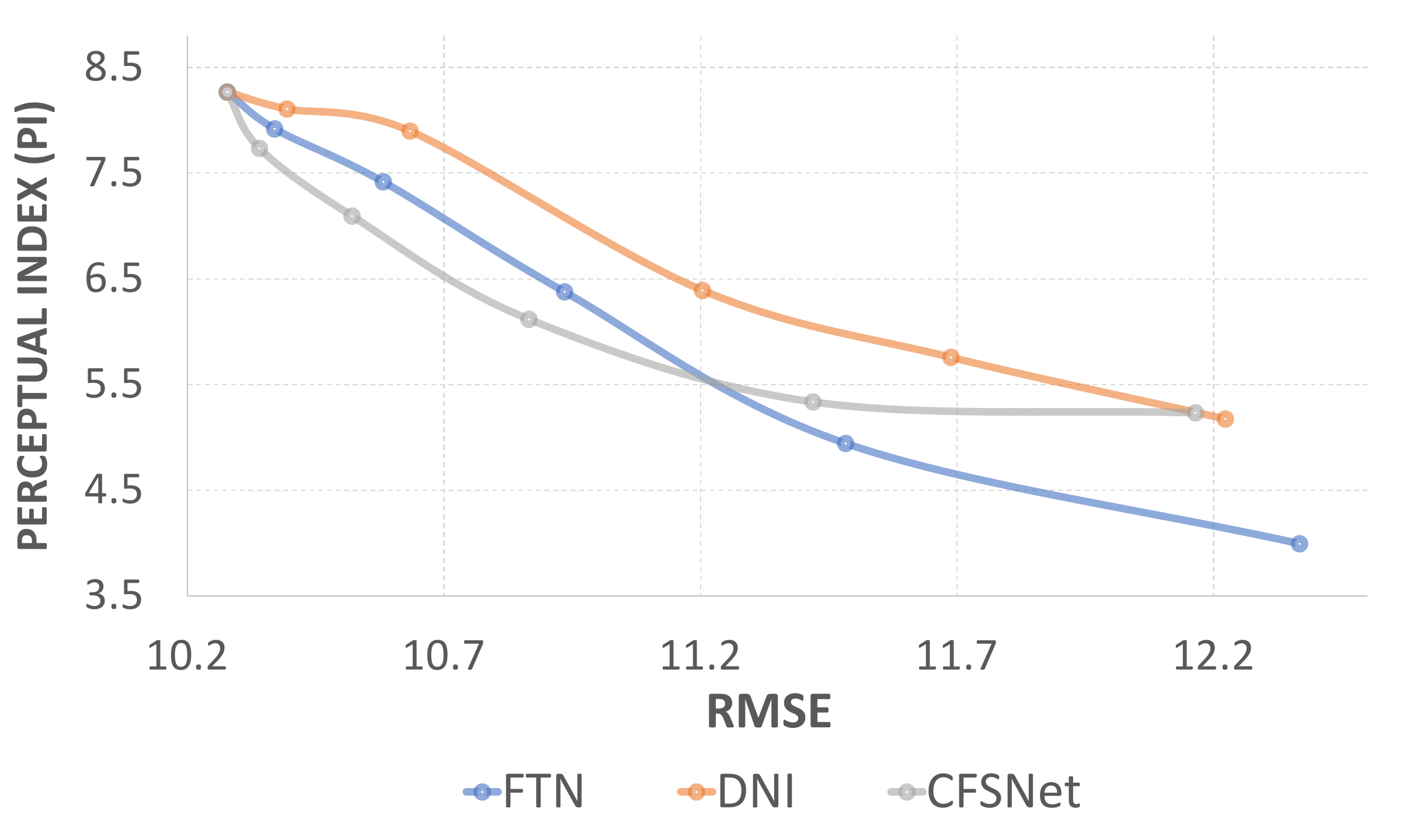}}\
		\subfloat[{SSIM-NIQE}]
		{\includegraphics[width=0.450\linewidth]{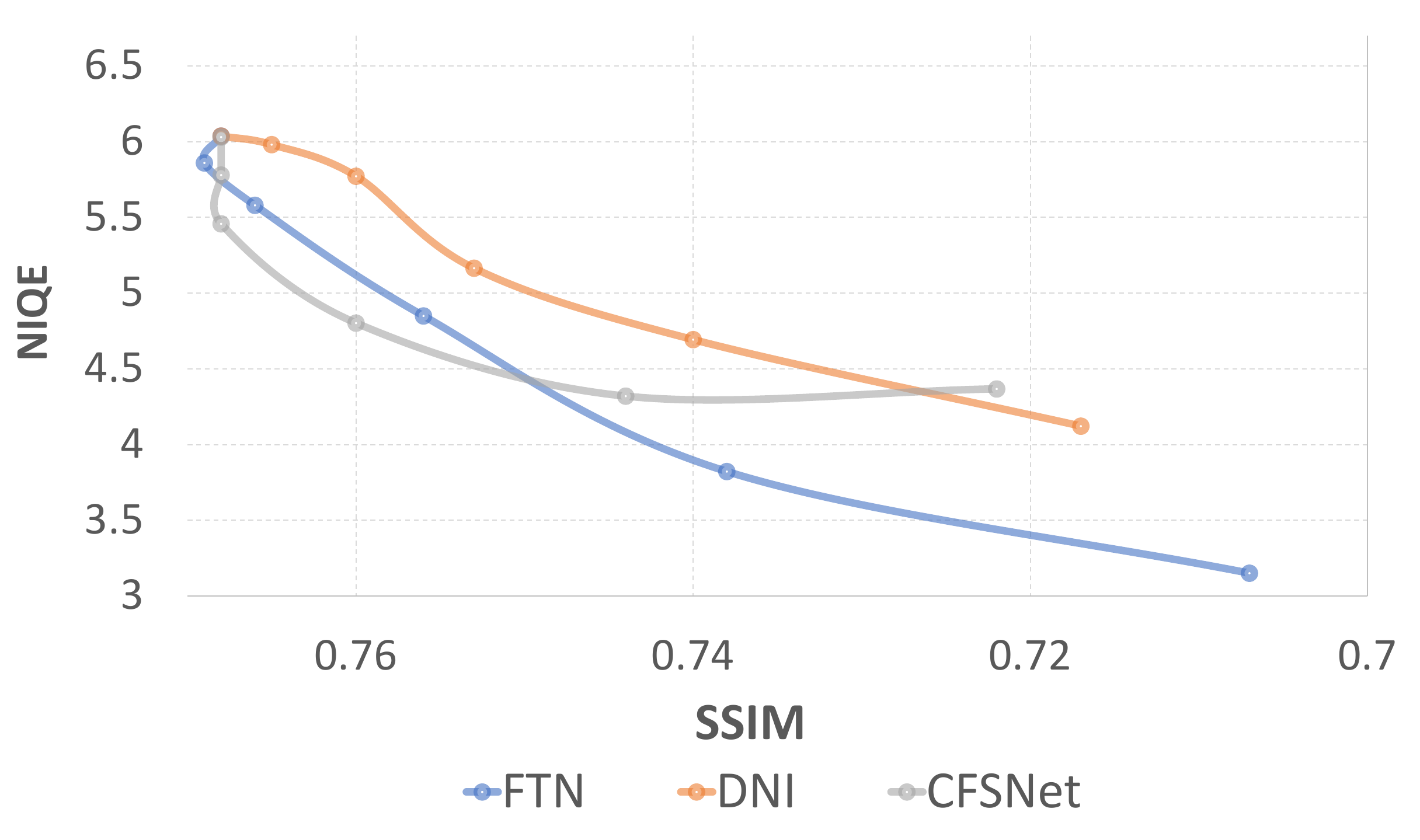}}\
		\caption{\small{\textbf{Results of PD-controllable image super-resolution ($\times$4).} Combined, various adaptation results, and results images are described in supplementary material.}}
		\label{fig:sr_results}
	\end{center}
\end{figure*}

\noindent\textbf{Comparison on performance.} First, for better interpolation, the performances on the intermediate levels compared to those of the network trained only for each level are important. In Table~\ref{tb:dn_result}, the results show that the short adaptation performance is similar or even better than the long one. In AdaFM-Net, FTN-gc4 and FTN-gc16 outperforms the other frameworks. In CFSNet-10, DNI outperforms other frameworks but the margin is not large. In CFSNet-10, the network is shallower than AdaFM-Net, which means that the parameter space can be easily linear. This can increase the performance of the linear interpolation (DNI). Compared to AdaFM and CFSNet, the interpolation performances of FTNs are superior to the others using much fewer computational costs.

\begin{figure*}[!t]
	\setlength{\belowcaptionskip}{-30pt}
	\begin{center}
		\captionsetup[subfigure]{labelformat=empty}
		\rotatebox{90}{\makebox[40mm][c]{{\textsc{FTN-gc16}}}}
		\subfloat
		{\includegraphics[width=0.440\linewidth]{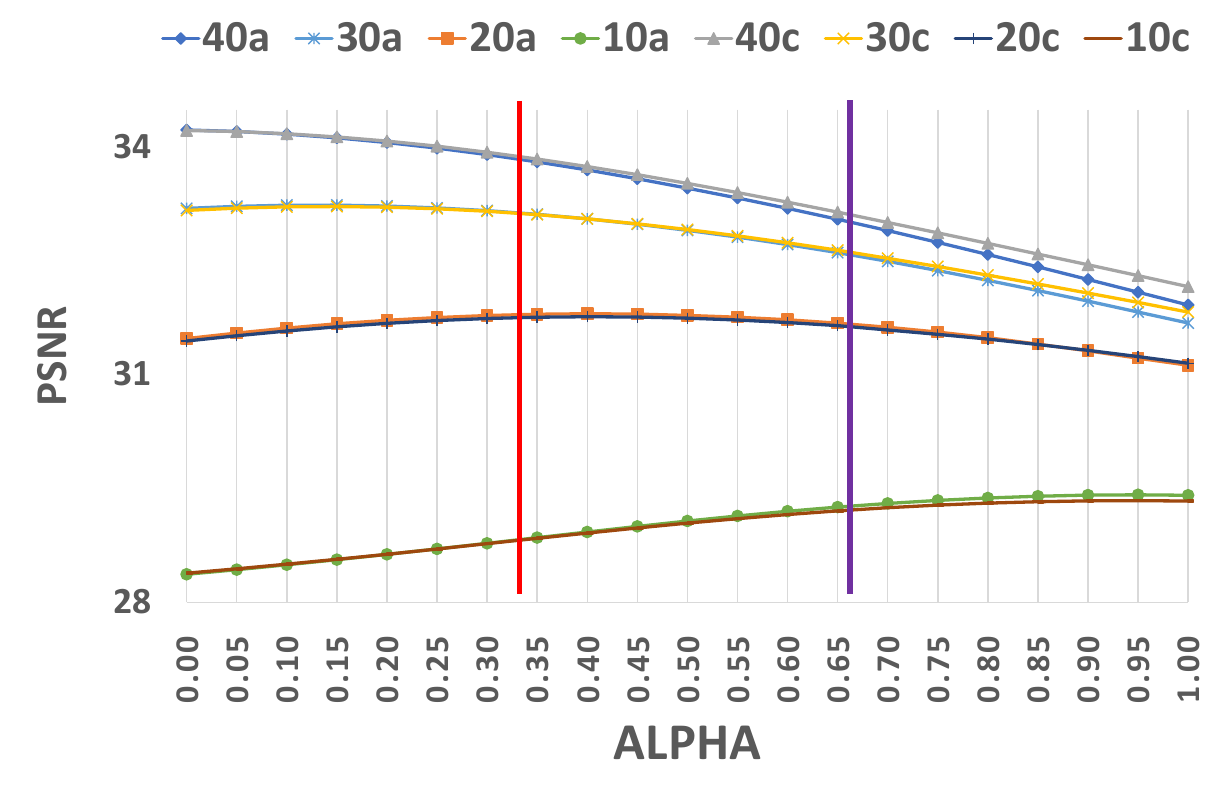}}\
		\rotatebox{90}{\makebox[40mm][c]{{\textsc{DNI}}}}
		\subfloat
		{\includegraphics[width=0.440\linewidth]{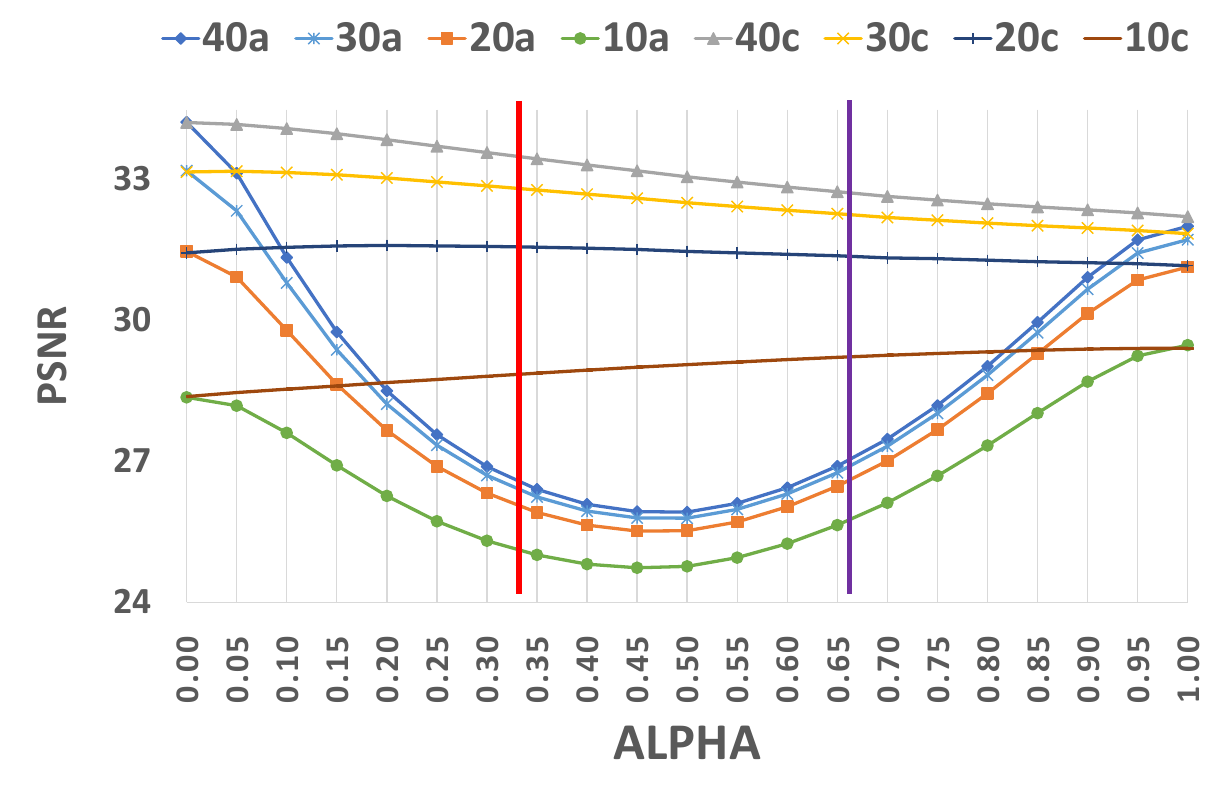}}\
		\caption{\small{\textbf{Smoothness analysis for deJPEG. ($q=40$ to $q=10$)} We plot \first{\textbf{$q=30$}} and \third{\textbf{$q=20$}} lines as linearly optimal interpolation points. Number indicates input quality factor, $a$ denotes AdaFM-Net network and $c$ denotes CFSNet-10 network}}
		\label{fig:graph_dj_network}
	\end{center}
\end{figure*}

However, compared to denoising task, the DNI shows different pattern in Fig.~\ref{fig:sr_results},  Fig.~\ref{fig:graph_dj_network} and Fig.~\ref{fig:dn_results}. Although DNI performs well for both end levels, it shows significantly unstable and low performance for the intermediate levels. This is because the fine-tuning process of DNI is just updating the parameters, without considering the initial state. Therefore, the relation between the parameters for the two levels gets weaker, and the interpolated parameters start not to behave as intended. From the intermediate images (in supplementary material and in Fig.~\ref{fig:dn_results}) for DNI, a little color difference can cause huge pixel-error (RMSE), but similar value in SSIM metric. On the other hand, FTN always takes the initial filters as input and also can be regularized by group convolutions. Therefore we can get the better interpolation performance. Our results are slightly inferior to CFSNet. However, our FTNs require extremely low computations and parameters.

\begin{figure*}[!t]
	\setlength{\belowcaptionskip}{-20pt}
	\begin{center}
		\captionsetup[subfigure]{labelformat=empty}
		\rotatebox{90}{\makebox[40mm][c]{{\textsc{FTN-gc16}}}}
		\subfloat
		{\includegraphics[width=0.440\linewidth]{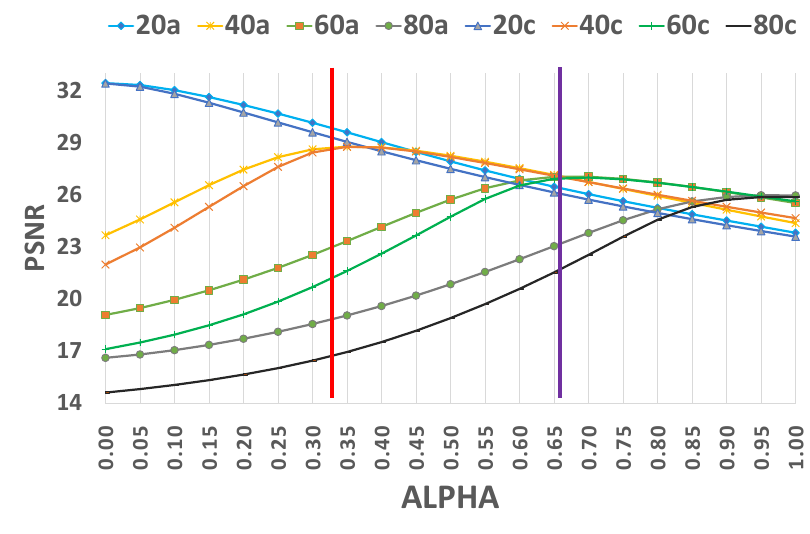}}\
		\rotatebox{90}{\makebox[40mm][c]{{\textsc{CFSNet}}}}
		\subfloat
		{\includegraphics[width=0.440\linewidth]{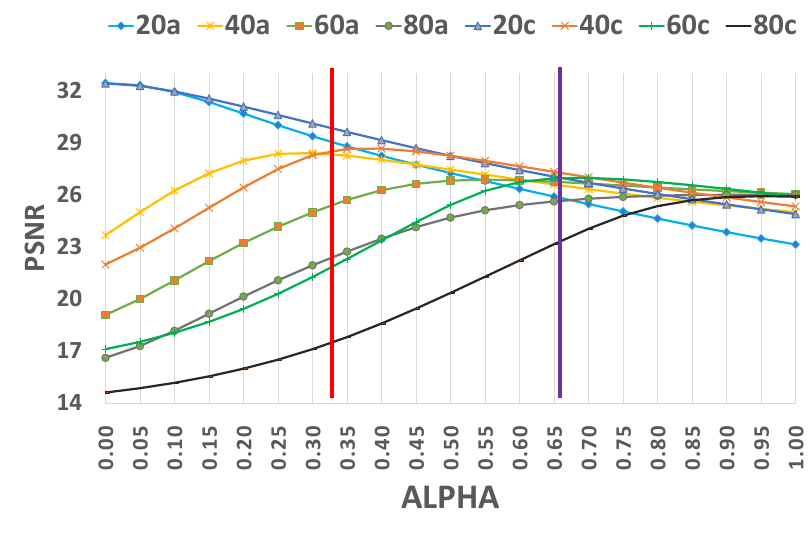}}\
		\\[-5ex]
		\rotatebox{90}{\makebox[40mm][c]{{\textsc{AdaFM}}}}
		\subfloat
		{\includegraphics[width=0.440\linewidth]{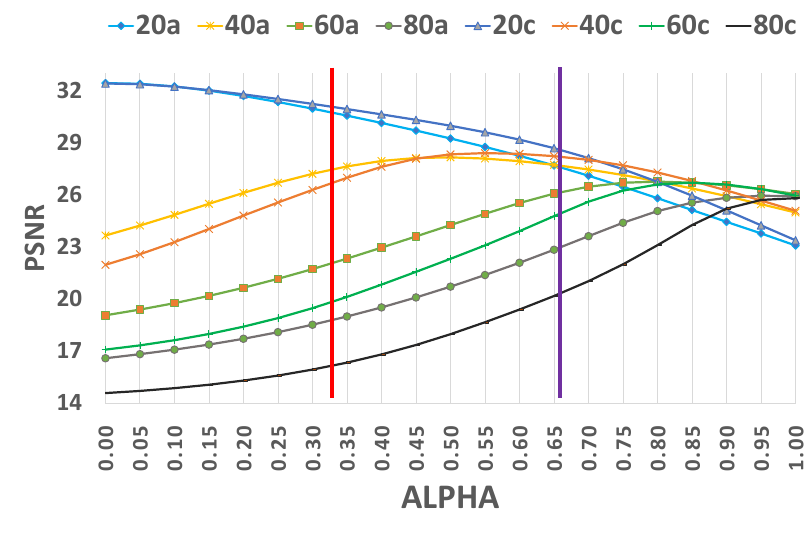}}\
		\rotatebox{90}{\makebox[40mm][c]{{\textsc{DNI}}}}
		\subfloat
		{\includegraphics[width=0.440\linewidth]{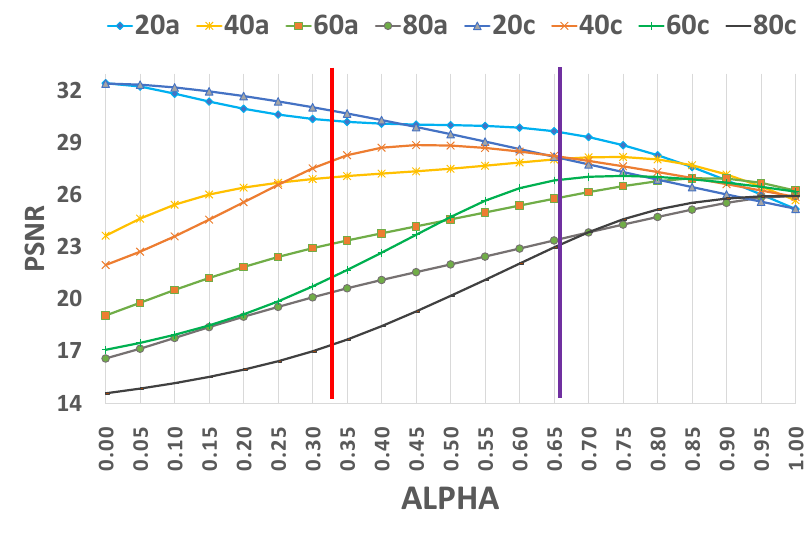}}\
		\caption{\small{\textbf{Smoothness analysis for denoising. ($\sigma=20$ to $\sigma=80$)} We plot \first{\textbf{$\sigma=40$}} and \third{\textbf{$\sigma=60$}} lines as linearly optimal interpolation points. Number indicates input quality factor, \textbf{$a$} denotes AdaFM-Net network and \textbf{$c$} denotes CFSNet-10 network. Our FTN-gc16 results show that the choice of $\alpha$ is closest to the lines}}
		\label{fig:graph_dn_network}
	\end{center}
\end{figure*}

\noindent\textbf{Interpretability.} For practical use, it will be essential for the users to know which value of $\alpha$ corresponds to which level. For example, in denoising task, suppose that we train a network to work between the levels $\sigma=20$ and $\sigma=80$. When we set $\alpha=0.5$, it is reasonable that the network will perform best for the level $\sigma=50$, which is the middle point of the interval. In other words, $\alpha$ has to be linear along with the level. Fig.~\ref{fig:graph_dn_network} shows the result of denoising task over various noise level $\sigma$ of the test set and the parameter $\alpha$. According to the figure, the maximum performance point for $\sigma=40$ and $\sigma=60$ best matches to the vertical lines of $\alpha=0.33$ and $\alpha=0.66$, compared to the other methods.

\begin{figure*}[!t]
	\setlength{\abovecaptionskip}{-1pt}
	\setlength{\belowcaptionskip}{-30pt}
	\begin{center}
		\captionsetup[subfigure]{labelformat=empty}
		\rotatebox{90}{\makebox[30mm][c]{\small{\textsc{Input}}}}
		\subfloat
		{\includegraphics[width=0.158\linewidth]{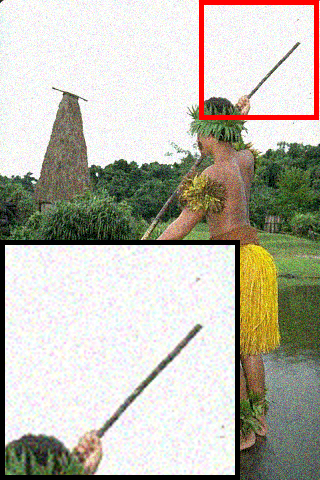}}\
		\hfill
		\rotatebox{90}{\makebox[30mm][c]{\small{\textsc{Clean}}}}
		\subfloat
		{\includegraphics[width=0.158\linewidth]{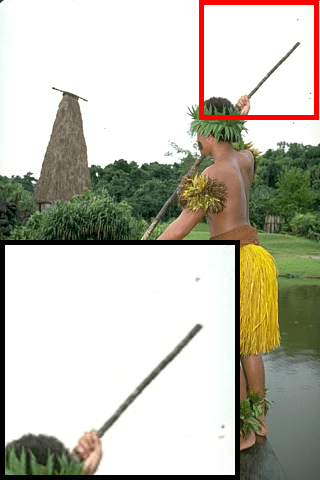}}\
		\\[-2ex]
		\rotatebox{90}{\makebox[20mm][c]{\small{\textsc{FTN-gc16}}}}
		\subfloat
		{\includegraphics[width=0.155\linewidth]{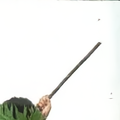}}\
		\hfill
		\subfloat
		{\includegraphics[width=0.155\linewidth]{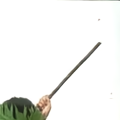}}\
		\hfill		
		\subfloat
		{\includegraphics[width=0.155\linewidth]{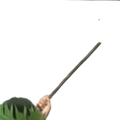}}\
		\hfill
		\subfloat
		{\includegraphics[width=0.155\linewidth]{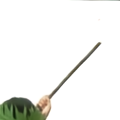}}\
		\hfill
		\subfloat
		{\includegraphics[width=0.155\linewidth]{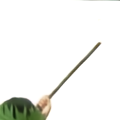}}\
		\hfill
		\subfloat
		{\includegraphics[width=0.155\linewidth]{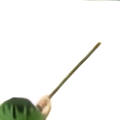}}\
		\\[-2ex]
		\rotatebox{90}{\makebox[20mm][c]{\small{\textsc{DNI}}}}
		\subfloat
		{\includegraphics[width=0.155\linewidth]{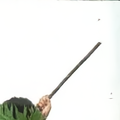}}\
		\hfill
		\subfloat
		{\includegraphics[width=0.155\linewidth]{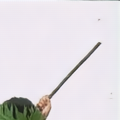}}\
		\hfill		
		\subfloat
		{\includegraphics[width=0.155\linewidth]{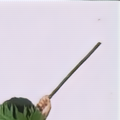}}\
		\hfill
		\subfloat
		{\includegraphics[width=0.155\linewidth]{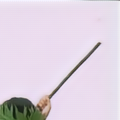}}\
		\hfill
		\subfloat
		{\includegraphics[width=0.155\linewidth]{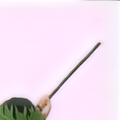}}\
		\hfill
		\subfloat
		{\includegraphics[width=0.155\linewidth]{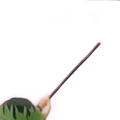}}\
		\\[-2ex]
		\rotatebox{90}{\makebox[20mm][c]{\small{\textsc{CFSNet}}}}
		\subfloat[$\alpha=0.0$]
		{\includegraphics[width=0.155\linewidth]{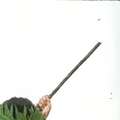}}\
		\hfill
		\subfloat[$\alpha=0.2$]
		{\includegraphics[width=0.155\linewidth]{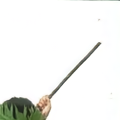}}\
		\hfill		
		\subfloat[$\alpha=0.4$]
		{\includegraphics[width=0.155\linewidth]{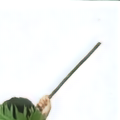}}\
		\hfill
		\subfloat[$\alpha=0.6$]
		{\includegraphics[width=0.155\linewidth]{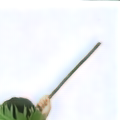}}\
		\hfill
		\subfloat[$\alpha=0.8$]
		{\includegraphics[width=0.155\linewidth]{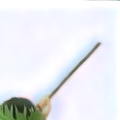}}\
		\hfill
		\subfloat[$\alpha=1.0$]
		{\includegraphics[width=0.155\linewidth]{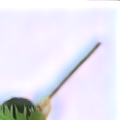}}\
		\caption{\small{\textbf{Denoising results on weak noise level ($\sigma=20$). When user control to the large $\alpha$, oversmoothing color artifacts are arose}}}
		\label{fig:dn_results}
	\end{center}
\end{figure*}

\noindent\textbf{Oversmoothing Artifacts.} In real-world applications, since the user may not know the degradation level, user hope to control the \textit{strength} of the denoising. We describe our visual denoising result on an extreme case in Fig. \ref{fig:dn_results}. In Fig. \ref{fig:dn_results}, the input noise level is 20, which means the optimal results come from $\alpha=0$ in all frameworks. $\alpha=1$, which is optimal for noise level 80, can over-smoothen the image. When $\alpha$ increases, DNI and CFSNet results show striking color artifacts in the background, while the FTN-gc16 results is much clean, which can be strength for real-world feedback-based systems. Since CFSNet exploits dual network structure and each network cannot consider the other level in the test phase. In contrast, in FTNs, two filters for both side of FTNs are correlated and regularized.

\subsection{Efficient Pixel-Adaptive Continuous Control}
\label{pixeladaptive}

\begin{figure}[!t]
	\setlength{\abovecaptionskip}{-5pt}
	\begin{center}
		\includegraphics[width=0.6\columnwidth]{"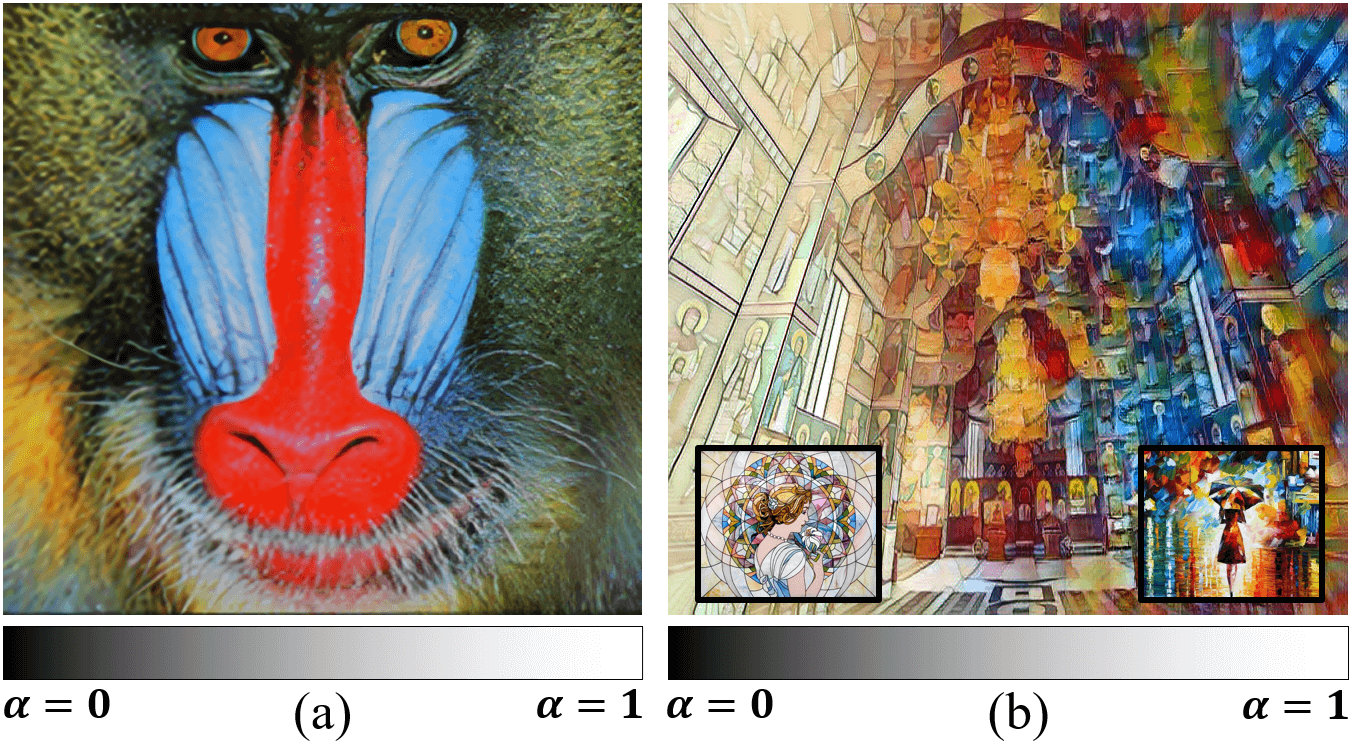}
	\end{center}
	\caption{\small{\textbf{Efficient pixel-adaptive control results.} Please zoom in for best view}}
	\label{fig:pixel_adaptive}
\end{figure}

Considering real-world imaging applications, the user wants to control not only the global level but also locally (pixel-wise). In this case, every pixel has its own imaging levels from $\alpha=0$ to $\alpha=1$. Naive pixel-adaptive controlling requires filters for every levels which can cause large memory issue. For efficient inference of pixel-adaptive continuous control, we propose a simple modification from the pixel-adaptive convolution. This is described as
\begin{equation}
\begin{aligned}
\textbf{Y} {} & = \textbf{X} {*}_{i,j} (\textbf{f} \times (1-{\alpha}_{i, j}) + FTN(\textbf{f}) \times {\alpha}_{i, j} ) \\
& = (\textbf{1}-{\textbf{A}}) \odot (\textbf{X} * \textbf{f})  + {\textbf{A}} \odot (\textbf{X} * FTN(\textbf{f})  )
\end{aligned}
\end{equation}

\noindent where $\textbf{f}$ is the global filter, $\textbf{*}_{i, j}$ is the pixel-adaptive convolution, ${\alpha}_{i, j}$ is the per-pixel level, and \textbf{A} is the global level-map that describes the pixel-wise levels. $\odot$ denotes element-wise multiplication. This modification makes implementation much simpler because only two global convolutions and multiplications are needed for pixel-adaptive control.
Examples are shown in Fig. \ref{fig:pixel_adaptive}. We test on two examples: \textit{PD-control} and \textit{style control}. In Fig. \ref{fig:pixel_adaptive} (a), from leftmost pixels to rightmost pixels, the PSNR decreases and the texture becomes sharper (higher perceptual quality) in continuously and in Fig. \ref{fig:pixel_adaptive} (b), the pixels are smoothly stylized from one style to the other. More results with high-resolution sources can be found in supplementary material.

\section{Conclusion}

In this paper, we define three conditions for the general and stable CLL framework: good adaptation, good interpolation and efficiency. To achieve these conditions, we propose Filter Transition Networks (FTNs) and stable initilization method. Non-linear structure of FTNs satisfies adaptation condition and regularized structure makes the interpolation smoothly. FTNs are extremely lightweight because of its data-agnostic structure. Results on general imaging tasks show that FTNs are better than the other unstable frameworks, and comparable to the other complex ones.

\par\vfill\par

\clearpage
%
%
\bibliographystyle{splncs04}
\bibliography{egbib}

\begin{thebibliography}{10}
\providecommand{\url}[1]{\texttt{#1}}
\providecommand{\urlprefix}{URL }
\providecommand{\doi}[1]{https://doi.org/#1}

\bibitem{DIV2k}
Agustsson, E., Timofte, R.: Ntire 2017 challenge on single image
  super-resolution: Dataset and study. In: The IEEE Conference on Computer
  Vision and Pattern Recognition (CVPR) Workshops (July 2017)

\bibitem{pirm2018}
Blau, Y., Mechrez, R., Timofte, R., Michaeli, T., Zelnik-Manor, L.: The 2018
  pirm challenge on perceptual image super-resolution. In: Proceedings of the
  European Conference on Computer Vision (ECCV). pp.~0--0 (2018)

\bibitem{pdt2018}
Blau, Y., Michaeli, T.: The perception-distortion tradeoff. In: Proceedings of
  the IEEE Conference on Computer Vision and Pattern Recognition. pp.
  6228--6237 (2018)

\bibitem{arcnn2015}
Dong, C., Deng, Y., Change~Loy, C., Tang, X.: Compression artifacts reduction
  by a deep convolutional network. In: Proceedings of the IEEE International
  Conference on Computer Vision. pp. 576--584 (2015)

\bibitem{srcnn2015}
Dong, C., Loy, C.C., He, K., Tang, X.: Image super-resolution using deep
  convolutional networks. IEEE transactions on pattern analysis and machine
  intelligence  \textbf{38}(2),  295--307 (2015)

\bibitem{argan2017}
Galteri, L., Seidenari, L., Bertini, M., Del~Bimbo, A.: Deep generative
  adversarial compression artifact removal. In: Proceedings of the IEEE
  International Conference on Computer Vision. pp. 4826--4835 (2017)

\bibitem{styletransfer2015}
Gatys, L.A., Ecker, A.S., Bethge, M.: A neural algorithm of artistic style.
  arXiv preprint arXiv:1508.06576  (2015)

\bibitem{gatys2017controlling}
Gatys, L.A., Ecker, A.S., Bethge, M., Hertzmann, A., Shechtman, E.: Controlling
  perceptual factors in neural style transfer. In: Proceedings of the IEEE
  Conference on Computer Vision and Pattern Recognition. pp. 3985--3993 (2017)

\bibitem{xavier2010}
Glorot, X., Bengio, Y.: Understanding the difficulty of training deep
  feedforward neural networks. In: Proceedings of the thirteenth international
  conference on artificial intelligence and statistics. pp. 249--256 (2010)

\bibitem{gan2014}
Goodfellow, I., Pouget-Abadie, J., Mirza, M., Xu, B., Warde-Farley, D., Ozair,
  S., Courville, A., Bengio, Y.: Generative adversarial nets. In: Advances in
  neural information processing systems. pp. 2672--2680 (2014)

\bibitem{guo2019toward}
Guo, S., Yan, Z., Zhang, K., Zuo, W., Zhang, L.: Toward convolutional blind
  denoising of real photographs. In: Proceedings of the IEEE Conference on
  Computer Vision and Pattern Recognition. pp. 1712--1722 (2019)

\bibitem{adafm2019}
He, J., Dong, C., Qiao, Y.: Modulating image restoration with continual levels
  via adaptive feature modification layers. In: Proceedings of the IEEE
  Conference on Computer Vision and Pattern Recognition. pp. 11056--11064
  (2019)

\bibitem{prelu2015}
He, K., Zhang, X., Ren, S., Sun, J.: Delving deep into rectifiers: Surpassing
  human-level performance on imagenet classification. In: Proceedings of the
  IEEE international conference on computer vision. pp. 1026--1034 (2015)

\bibitem{huang2017arbitrary}
Huang, X., Belongie, S.: Arbitrary style transfer in real-time with adaptive
  instance normalization. In: Proceedings of the IEEE International Conference
  on Computer Vision. pp. 1501--1510 (2017)

\bibitem{perceptual2016}
Johnson, J., Alahi, A., Fei-Fei, L.: Perceptual losses for real-time style
  transfer and super-resolution. In: European conference on computer vision.
  pp. 694--711. Springer (2016)

\bibitem{vdsr2016}
Kim, J., Kwon~Lee, J., Mu~Lee, K.: Accurate image super-resolution using very
  deep convolutional networks. In: Proceedings of the IEEE conference on
  computer vision and pattern recognition. pp. 1646--1654 (2016)

\bibitem{alexnet}
Krizhevsky, A., Sutskever, I., Hinton, G.E.: Imagenet classification with deep
  convolutional neural networks. In: Advances in neural information processing
  systems. pp. 1097--1105 (2012)

\bibitem{lapsrn2017}
Lai, W.S., Huang, J.B., Ahuja, N., Yang, M.H.: Deep laplacian pyramid networks
  for fast and accurate super-resolution. In: Proceedings of the IEEE
  conference on computer vision and pattern recognition. pp. 624--632 (2017)

\bibitem{srgan2017}
Ledig, C., Theis, L., Husz{\'a}r, F., Caballero, J., Cunningham, A., Acosta,
  A., Aitken, A., Tejani, A., Totz, J., Wang, Z., et~al.: Photo-realistic
  single image super-resolution using a generative adversarial network. In:
  Proceedings of the IEEE conference on computer vision and pattern
  recognition. pp. 4681--4690 (2017)

\bibitem{edsr2017}
Lim, B., Son, S., Kim, H., Nah, S., Mu~Lee, K.: Enhanced deep residual networks
  for single image super-resolution. In: Proceedings of the IEEE conference on
  computer vision and pattern recognition workshops. pp. 136--144 (2017)

\bibitem{cocodataset}
Lin, T.Y., Maire, M., Belongie, S., Hays, J., Perona, P., Ramanan, D.,
  Doll{\'a}r, P., Zitnick, C.L.: Microsoft coco: Common objects in context. In:
  European conference on computer vision. pp. 740--755. Springer (2014)

\bibitem{nlrn2018}
Liu, D., Wen, B., Fan, Y., Loy, C.C., Huang, T.S.: Non-local recurrent network
  for image restoration. In: Advances in Neural Information Processing Systems.
  pp. 1673--1682 (2018)

\bibitem{cbsd68}
Martin, D., Fowlkes, C., Tal, D., Malik, J., et~al.: A database of human
  segmented natural images and its application to evaluating segmentation
  algorithms and measuring ecological statistics. Iccv Vancouver: (2001)

\bibitem{mildenhall2018burst}
Mildenhall, B., Barron, J.T., Chen, J., Sharlet, D., Ng, R., Carroll, R.: Burst
  denoising with kernel prediction networks. In: Proceedings of the IEEE
  Conference on Computer Vision and Pattern Recognition. pp. 2502--2510 (2018)

\bibitem{cgan}
Mirza, M., Osindero, S.: Conditional generative adversarial nets. arXiv
  preprint arXiv:1411.1784  (2014)

\bibitem{niqe}
Mittal, A., Soundararajan, R., Bovik, A.C.: Making a “completely blind”
  image quality analyzer. IEEE Signal Processing Letters  \textbf{20}(3),
  209--212 (2012)

\bibitem{live1}
Moorthy, A.K., Bovik, A.C.: Visual importance pooling for image quality
  assessment. IEEE journal of selected topics in signal processing
  \textbf{3}(2),  193--201 (2009)

\bibitem{dcgan}
Radford, A., Metz, L., Chintala, S.: Unsupervised representation learning with
  deep convolutional generative adversarial networks. arXiv preprint
  arXiv:1511.06434  (2015)

\bibitem{sheng2018avatar}
Sheng, L., Lin, Z., Shao, J., Wang, X.: Avatar-net: Multi-scale zero-shot style
  transfer by feature decoration. In: Proceedings of the IEEE Conference on
  Computer Vision and Pattern Recognition. pp. 8242--8250 (2018)

\bibitem{dynamicnet2019}
Shoshan, A., Mechrez, R., Zelnik-Manor, L.: Dynamic-net: Tuning the objective
  without re-training for synthesis tasks. In: Proceedings of the IEEE
  International Conference on Computer Vision. pp. 3215--3223 (2019)

\bibitem{instance_norm}
Ulyanov, D., Vedaldi, A., Lempitsky, V.: Instance normalization: The missing
  ingredient for fast stylization. arXiv preprint arXiv:1607.08022  (2016)

\bibitem{cfsnet2019}
Wang, W., Guo, R., Tian, Y., Yang, W.: Cfsnet: Toward a controllable feature
  space for image restoration. In: Proceedings of the IEEE International
  Conference on Computer Vision. pp. 4140--4149 (2019)

\bibitem{dni2019}
Wang, X., Yu, K., Dong, C., Tang, X., Loy, C.C.: Deep network interpolation for
  continuous imagery effect transition. In: Proceedings of the IEEE Conference
  on Computer Vision and Pattern Recognition. pp. 1692--1701 (2019)

\bibitem{esrgan2018}
Wang, X., Yu, K., Wu, S., Gu, J., Liu, Y., Dong, C., Qiao, Y., Change~Loy, C.:
  Esrgan: Enhanced super-resolution generative adversarial networks. In:
  Proceedings of the European Conference on Computer Vision (ECCV). pp.~0--0
  (2018)

\bibitem{ssim}
Wang, Z., Bovik, A.C., Sheikh, H.R., Simoncelli, E.P., et~al.: Image quality
  assessment: from error visibility to structural similarity. IEEE transactions
  on image processing  \textbf{13}(4),  600--612 (2004)

\bibitem{resnext}
Xie, S., Girshick, R., Doll{\'a}r, P., Tu, Z., He, K.: Aggregated residual
  transformations for deep neural networks. In: Proceedings of the IEEE
  conference on computer vision and pattern recognition. pp. 1492--1500 (2017)

\bibitem{dncnn2017}
Zhang, K., Zuo, W., Chen, Y., Meng, D., Zhang, L.: Beyond a gaussian denoiser:
  Residual learning of deep cnn for image denoising. IEEE Transactions on Image
  Processing  \textbf{26}(7),  3142--3155 (2017)

\bibitem{zhang2018ffdnet}
Zhang, K., Zuo, W., Zhang, L.: Ffdnet: Toward a fast and flexible solution for
  cnn-based image denoising. IEEE Transactions on Image Processing
  \textbf{27}(9),  4608--4622 (2018)

\bibitem{rcan2018}
Zhang, Y., Li, K., Li, K., Wang, L., Zhong, B., Fu, Y.: Image super-resolution
  using very deep residual channel attention networks. In: Proceedings of the
  European Conference on Computer Vision (ECCV). pp. 286--301 (2018)

\bibitem{rdn2018}
Zhang, Y., Tian, Y., Kong, Y., Zhong, B., Fu, Y.: Residual dense network for
  image super-resolution. In: Proceedings of the IEEE Conference on Computer
  Vision and Pattern Recognition. pp. 2472--2481 (2018)

\end{thebibliography}
\end{document}